\documentclass[10pt,journal,compsoc]{IEEEtran}

\usepackage{caption}

\ifCLASSOPTIONcompsoc
  \usepackage[nocompress]{cite}
\else
  \usepackage{cite}
\fi
\ifCLASSINFOpdf
  \usepackage[pdftex]{graphicx}
  \DeclareGraphicsExtensions{.pdf,.jpeg,.png}
\else
\fi

\usepackage{url}

\usepackage{amsmath}
\usepackage{amsfonts}
\usepackage{booktabs} 
\usepackage{epstopdf}
\usepackage{multirow}
\usepackage[section]{placeins}
\usepackage{subfigure}
\newcommand{\loss}{\mathcal{L}}

\newcommand{\actfn}{\phi}

\newcommand{\R}{\mathbf{R}}

\newcommand{\W}{\mathbf{W}}
\newcommand{\bias}{\mathbf{b}}
\newcommand{\act}{\mathbf{z}}

\newcommand{\Ri}{\mathcal{R}_i}
\newcommand{\Rj}{\mathcal{R}_j}

\newcommand{\uu}{\mathbf{u}}
\newcommand{\ut}{\tilde{\mathbf{u}}}
\newcommand{\uh}{\hat{\mathbf{u}}}

\newcommand{\ui}{\mathbf{u}_i}
\newcommand{\vj}{\mathbf{v}_j}
\newcommand{\uti}{\tilde{\mathbf{u}}_i}
\newcommand{\uhi}{\hat{\mathbf{u}}_i}
\newcommand{\vtj}{\tilde{\mathbf{v}}_j}
\newcommand{\vhj}{\hat{\mathbf{v}}_j}

\newcommand{\mypar}[1]{\vspace{0.05in} \noindent \textbf{#1 \,}}

\usepackage{mathrsfs}
\usepackage{algorithmicx}
\usepackage[ruled]{algorithm}
\usepackage[noend]{algpseudocode}

\usepackage{indentfirst}

\addtocounter{footnote}{0}

\begin{document}
%
\title{\vspace{-10pt}Neural Collaborative Autoencoder}

%
%
%

%
%

\author{Qibing Li, Xiaolin Zheng,~\IEEEmembership{Senior Member,~IEEE,} Xinyue Wu

\IEEEcompsocitemizethanks{\IEEEcompsocthanksitem Q. Li, X. Zheng and X. Wu are with the College of Computer Science and Technology, Zhejiang university, Hangzhou, China.\protect\\
E-mail: qblee@zju.edu.cn, xlzheng@zju.edu.cn and wxinyue@zju.edu.cn
}
\thanks{Manuscript received April 19, 2005; revised August 26, 2015.}}

%
%

\markboth{IEEE TRANSACTIONS ON KNOWLEDGE AND DATA ENGINEERING (under review)}%
{Shell \MakeLowercase{\textit{et al.}}: Bare Advanced Demo of IEEEtran.cls for IEEE Computer Society Journals}
%



\IEEEtitleabstractindextext{%
\begin{abstract}
\noindent In recent years, deep neural networks have yielded state-of-the-art performance on several tasks. Although some recent works have focused on combining deep learning with recommendation, we highlight three issues of existing models. First, these models cannot work on both explicit and implicit feedback, since the network structures are specially designed for one particular case. Second, due to the difficulty on training deep neural networks, existing explicit models do not fully exploit the expressive potential of deep learning. Third, neural network models are easier to overfit on the implicit setting than shallow models. To tackle these issues, we present a generic recommender framework called \textit{Neural Collaborative Autoencoder} (NCAE) to perform collaborative filtering, which works well for both explicit feedback and implicit feedback. NCAE can effectively capture the subtle hidden relationships between interactions via a non-linear matrix factorization process. To optimize the deep architecture of NCAE, we develop a three-stage pre-training mechanism that combines supervised and unsupervised feature learning. Moreover, to prevent overfitting on the implicit setting, we propose an error reweighting module and a sparsity-aware data-augmentation strategy. Extensive experiments on three real-world datasets demonstrate that NCAE can significantly advance the state-of-the-art. 





\end{abstract}

}

\maketitle

\IEEEdisplaynontitleabstractindextext

%
\IEEEpeerreviewmaketitle

\ifCLASSOPTIONcompsoc
\IEEEraisesectionheading{\section{Introduction}\label{sec:introduction}}
\else
\section{Introduction}
\label{sec:introduction}
\fi
\IEEEPARstart{I}{n} recent years, recommender systems (RS) have played a significant role in E-commerce services. A good recommender system may enhance both satisfaction for users and profit for content providers. For example, nearly 80\% of movies watched on Netflix are recommended by RS \cite{Gomez2016The}. The key to design such a system is to predict users' preference on items based on past activities, which is known as collaborative filtering (CF) \cite{sarwar2001item}. Among the various CF methods, matrix factorization (MF) \cite{hu2008collaborative,koren2008factorization,koren2009matrix,shi2014collaborative} is the most used one, which models the user-item interaction function as the inner product of user latent vector and item latent vector. Due to the effectiveness of MF, many integrated models have been devised, such as CTR \cite{wang2011collaborative}, HFT \cite{mcauley2013hidden} and timeSVD \cite{koren2010collaborative}. However, the inner product is not sufficient for capturing subtle hidden factors from the interaction data \cite{he2017neural}.

Currently, a trend in the recommendation literature is the utilization of deep learning to handle the auxiliary information \cite{wang2015collaborative,li2015deep, wang2016crae} or directly model the interaction function \cite{sedhain2015autorec,strub2016hybrid,wu2016collaborative,he2017neural}. Thus, based on these two usage scenarios, deep learning based recommender systems can be roughly categorized into integration models and neural network models \cite{zhang2017deep}.  Integration models utilize deep neural networks to extract the hidden features of auxiliary information, such as item descriptions \cite{wang2015collaborative,wang2016crae}, user profiles \cite{li2015deep} and knowledge bases \cite{zhang2016collaborative}. The features are then integrated into the CF framework to perform hybrid recommendation.  Although integration models involve both deep learning and CF, they actually belong to MF-based models because they use an inner product to model the interaction data, and thus face the same issue like MF.

On the other hand, neural network models directly perform collaborative filtering via modeling the interaction data. Due to the effectiveness of deep components, neural network models are able to discover the non-linear hidden relationships from data \cite{lecun2015deep, strub2016hybrid}. For example, Collaborative Filtering Network (CFN) \cite{strub2016hybrid} is a state-of-the-art model for explicit feedback, which utilizes DAE \cite{vincent2008extracting} to encode sparse user/item preferences (rows or columns of the observed rating matrix) and aims to reconstruct them in the decoder layer. However, we notice that existing explicit neural models do not fully exploit the representation power of deep architectures; stacking more layers yields little performance gain \cite{zheng2016neural,strub2016hybrid}. This is mainly caused by two reasons. First, without a proper pre-training strategy, training deep neural networks is difficult \cite{glorot2010understanding}. Second, due to the sparse nature of RS, conventional layer-wise unsupervised pre-training \cite{hinton2006reducing,bengio2007greedy} does not work in this case\footnote{Supervised pre-training in the first hidden layer is critical to the performance, since unsupervised reconstruction method may lose user/item information. (see Sec. \ref{sec:experi_explict})}. Besides, neural network models are easier to overfit on the implicit setting due to the highly non-linear expressiveness (i.e., predicting all ratings as 1, since observed interactions are all converted as 1). Neural Collaborative Filtering (NCF) \cite{he2017neural} is a state-of-the-art implicit neural model that can capture the non-linear hidden factors while combatting overfitting (NCF samples negative feedback from unobserved data to perform pairwise learning like BPR). However, the NCF architecture is designed at the interaction level, which is time-consuming to rank all items for all users during evaluation. Furthermore, providing more item correlation patterns may be the key factors to combat overfitting for implicit neural models (see the discussion in Sec. \ref{sec:sparsity-aware}).


To address the aforementioned issues, we present a simple deep learning based recommender framework called \textit{Neural Collaborative Autoencoder} (NCAE) for both explicit feedback and implicit feedback. The central idea of NCAE is to learn hidden structures that can reconstruct user/item preferences via a non-linear matrix factorization process. The NCAE architecture is designed at the user/item level, which takes the sparse user or item vectors as inputs for batch training and evaluation. By utilizing a sparse forward module and a sparse backward module, NCAE is scalable to large datasets and robust to sparse data; a new training loss is also designed for sparse user/item inputs. We further develop a novel three-stage pre-training mechanism, combining supervised and unsupervised feature learning to train the deep architecture of NCAE. As a result, NCAE with deeper architectures is more powerful and expressive for explicit feedback. Besides, NCAE includes two ingredients to prevent overfitting on the implicit setting: an error reweighting module that treats all unobserved interactions as negative feedback (i.e., whole-based method \cite{hu2008collaborative, he2016fast}), and a sparsity-aware data-augmentation strategy that provides more item correlation patterns in training and discovers better ranking positions of true positive items during inference. The key contributions of this paper are summarized as follows: 
\begin{itemize}
\item We present a scalable and robust recommender framework named NCAE for both explicit feedback and implicit feedback, where we adapt several effective approaches from the deep learning literature to the recommendation domain, including autoencoders, dropout, pre-training and data-augmentation. NCAE can capture the subtle hidden factors via a non-linear matrix factorization process.
\item Thanks to the effectiveness of three-stage pre-training mechanism, NCAE can exploit the representation power of deep architectures to learn high-level abstractions, bringing better generalization. We further explore several variants of our proposed pre-training mechanism and compare our method with conventional pre-training strategies to better understand its effectiveness.
\item Compared with other implicit neural models, NCAE can utilize the error reweighting module and the sparsity-aware data augmentation to combat overfitting, and thus performs well on the implicit setting without negative sampling.
\item Extensive experiments on three real-world datasets demonstrate the effectiveness of NCAE on both explicit and implicit settings.
\end{itemize}

The paper is organized as follows. In Section \ref{sec:related_work}, we discuss related work on applying neural networks to recommender systems. In Section \ref{sec:problem}, we provide the problem definition and notations. We describe the proposed NCAE in Section \ref{sec:methodology}. We conduct experiments in Section \ref{sec:experiments} before concluding the paper in Section \ref{sec:conclusions}.

\section{Related Work}
\label{sec:related_work}
We review the existing models in two groups, including integration models and neural network models \cite{zhang2017deep}.

Integration models combine DNNs with the CF framework such as PMF \cite{mnih2008probabilistic} to perform hybrid recommendation, which can deal with the cold-start problem and alleviate data sparsity. The deep components of integration models are primarily used to extract the hidden features of auxiliary information \cite{wang2015collaborative,wang2016crae,li2015deep,zhang2016collaborative,cheng2016wide}. For example, Collaborative Deep Learning (CDL) \cite{wang2015collaborative} integrates Stack Denoising Autoencoder (SDAE) \cite{vincent2010stacked} and PMF \cite{mnih2008probabilistic} into a unified probabilistic graph model to jointly perform deep content feature learning and collaborative filtering. However, integration models utilize an inner product to model the interaction data, which is not sufficient for capture the complex structure of interaction data \cite{he2017neural}. Different from integration models, NCAE can model the interaction function via a non-linear matrix factorization process. Our focus is to design a deep learning based recommender framework for both explicit feedback and implicit feedback. Therefore, we remain the hybrid recommender (combining NCAE with deep content feature learning) to the future work.

Neural network models utilize DNNs to learn the interaction function from data, which are able to discover the non-linear hidden relationships \cite{lecun2015deep, strub2016hybrid, he2017neural}. We may further divide neural network models into two sub-categories: explicit neural models and implicit neural models. Restricted Boltzmann Machine (RBM) for CF \cite{salakhutdinov2007restricted} is the early pioneer work that applies neural networks to explicit feedback. Recently, autoencoders have become a popular building block for explicit models \cite{sedhain2015autorec,strub2016hybrid}. For example, \cite{strub2016hybrid} proposes CFN that achieves the best performance for explicit feedback (with well-tuned parameters and additional data pre-precessing). Compared to CFN, our NCAE employs a new training loss to balance the impact of sparse user/item vectors and achieves comparable performance without complex procedures of CFN. By utilizing the three-stage pre-training, NCAE with deeper architectures is more powerful and expressive than CFN. To our best knowledge, NCAE obtains a new state-of-the-art result for explicit feedback. On the other hand, implicit neural models also exploit the representation capacity of neural networks and combat overfitting by sampling negative interactions. Compared to the state-of-the-art implicit model NCF, our NCAE takes user vectors as inputs for batch evaluation; NCAE explicitly gathers information from other users for batch training, which may be a better network structure for CF (see Sec. \ref{sec:implicit_protocol}). Besides, the error reweighting module and the sparsity-aware data-augmentation can provide more item correlation patterns for NCAE, greatly enhancing the top-M recommendation performance. Overall, NCAE is a generic framework for CF that performs well on the two settings.

\section{Problem Definition}
\label{sec:problem}
We start by introducing some basic notations. Matrices are written as bold capital letters (e.g., $\mathbf{X}$) and vectors are expressed as bold lowercase letters (e.g., $\mathbf{x}$). We denote the $i$-th row of matrix $\mathbf{X}$ by $\mathbf{X}_i$ and its ($i$, $j$)-th element by $\mathbf{X}_{ij}$. For vectors, $\mathbf{x}_i$ denotes the $i$-th element of $\mathbf{x}$.

\begin{figure}[t]
\centering
\includegraphics[width=\columnwidth]{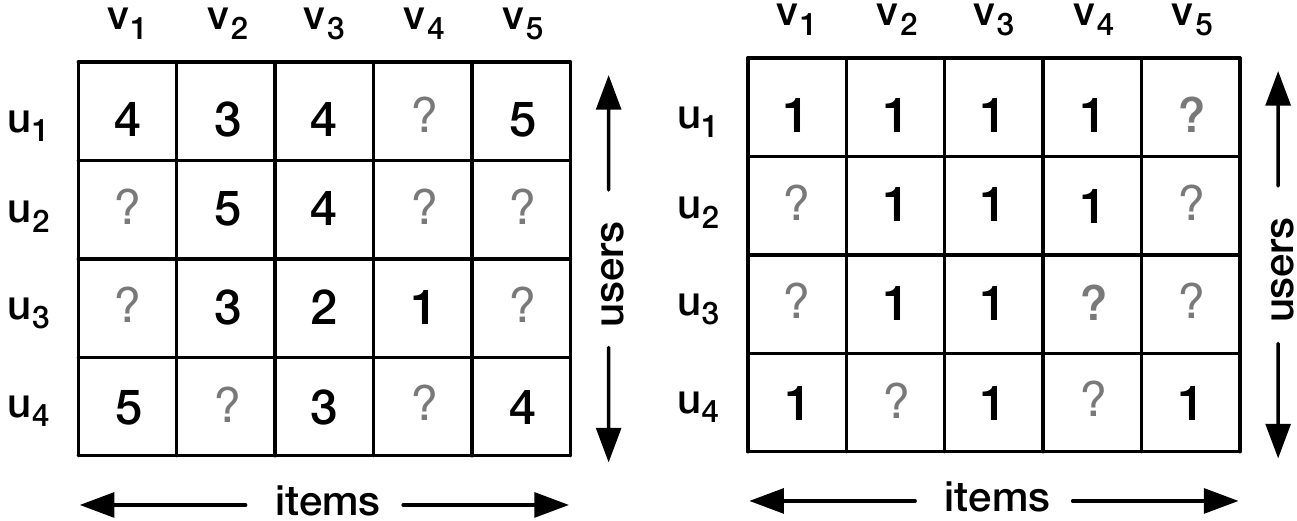}
\vspace{-3pt}
\caption{A simple example illustrating the rating matrices for explicit feedback and implicit feedback.}
\label{fig:intuition}
\end{figure}

In collaborative filtering, we have $M$ users, $N$  items and a partially observed user-item rating matrix $\R=[\R_{ij}]_{M \times N}$. Fig. \ref{fig:intuition} shows an example of explicit feedback and implicit feedback. Generally, explicit feedback problem can be regarded as a regression problem on observed ratings, while the implicit one is known as a one-class classification problem \cite{pan2008one} based on interactions.
 
We let $\mathcal{R}$ denote the set of user-item pairs $(i,j,r_{ij})$ where values are non-zero, $\bar{\mathcal{R}}$ denote the set of unobserved, missing triples. Let $\Ri$ denote the set of item preferences in the training set for a particular user $i$; similar notations for $\bar{\mathcal{R}}_i$. The goal of the recommender system is to pick a subset of items from the candidate set $\bar{\mathcal{R}}_i$ for each user $i$. Besides, we use a sparse vector $\ui \!\in\! \mathbb{R}^{N}$ as the input of NCAE, which has only $|\Ri|$ observed values of user $i$: $\uu_{ij}=r_{ij}$ if $j$ is in $\Ri$, otherwise, $\uu_{ij}=0$; $\uti$, $\uhi$ are the corrupted version of $\ui$ and the dense estimate of ${\R}_i$; similar notations $\Rj$, $\vj$, $\vtj$ and $\vhj$ for each item $j$. 

For neural network settings, $K_{l}$, $\W^l \!\in\! \mathbb{R}^{K_{l} \times K_{l-1}}$, $\bias^l \!\in\! \mathbb{R}^{K_{l}} $ are represented as the hidden dimension, the weight matrix and the bias vector of the $l$-th layer. $L$ is the number of layers. For convenience, we use $\W^+$ to denote the collection of all weights and biases. $\actfn^l$ and $\act^l \!\in\! \mathbb{R}^{K_{l}}$ are the activation function and the activation output vector of layer $l$, respectively. In our work, we use $\tanh(\cdot)$ as activation function $\actfn^l$ for every layer $l$.
\section{Proposed Methdology}
\label{sec:methodology}
In this section, we first introduce a new recommender framework --- Neural Collaborative Autoencoder (NCAE), which consists of an input dropout module, a sparse forward module, a sparse backward module and an error reweighing module. Then we propose a sparsity-aware data-augmentation strategy and a three-stage layer-wise pre-training mechanism to enhance the performance.

\subsection{Neural Collaborative Autoencoder}
\label{sec:dca}
\begin{figure*}
\centering
\includegraphics[width=18cm]{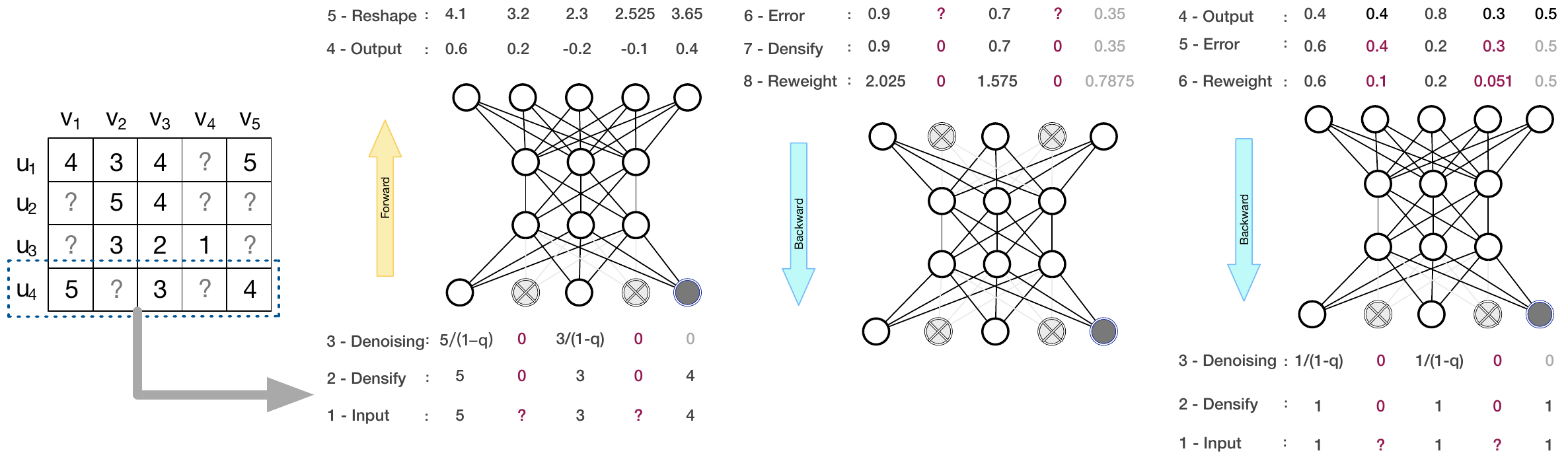}
\vspace{-3pt}
\caption{A three-layer NCAE. Left: Sparse forward procedure with input dropout. Middle: Sparse backward procedure for the explicit feedback, where errors for known values are reweighed by $2.25$ due to output reshape ($nn(\uti)=2.25*\act^L_i+2.75$.) Right: Sparse backward for implicit feedback, where popularity-based error reweighting is used on unknown values.}
\label{fig:dcae}
\end{figure*}
Previous work based on autoencoders can be simply categoried into two style: user-based and item-based \cite{strub2016hybrid,sedhain2015autorec,wu2016collaborative}. For simplicity, the user-based autoencoder is considered in our paper\footnote{Item-based autoencoders will be used for explicit feedback in the experiments.}. The collaborative filtering problem can be interpreted based on autoencoders: \textit{transform the sparse vector $\ui$ into dense vector $\uhi$ for every user $i$}. Formally, user-based NCAE is defined as:
\begin{equation}
\label{eq:fordward_out}
	\uhi=nn(\uti)=\act^L_i=\phi^L(\W^L(..\phi^1(\W^1\uti+ \bias^1)..)+\bias^L),
\end{equation}
where the first $L$-1 layer of NCAE aims at building a low-rank representation for user $i$, and the last layer $L$ can be regarded as item representation learning or a task-related decoding part. For the explicit setting, we make a slight modification and let $nn(\uti)=2.25*\act^L_i+2.75$, since ratings are 10 scale with 0.5 constant increments (0.5 to 5.0) and the range of $\tanh(\cdot)$ is [-1,1]. 

In general, NCAE is strongly linked with matrix factorization. For NCAE with only one hidden layer and no output activation function, $nn(\uti)=\W^2\phi^1(\W^1\uti+ \bias^1)+\bias^2$ can be reformulated as:
\begin{equation}
	\uhi=nn(\uti)=\underbrace{[\W^{2}\;\mathbf{I}_{N}]}_{\mathbf{V}})
	\underbrace{\left[\begin{array}{c}
    \phi^1(\W^1\uti+ \bias^1)\\ 
    \bias^2\\
  \end{array}\right]}_{\mathbf{U}_i},
\end{equation}
where $\mathbf{V}$ is the item latent matrix and $\mathbf{U_i}$ is the user latent factors for user $i$. NCAE with more hidden layers and output activation function $\phi^L$ can be regarded as \textit{non-linear matrix factorization}, aiming at learning deep representations. A nice benefit of the learned NCAE is that it can fill in every vector $\ui$, even if that vector was not in the training data.

Fig. \ref{fig:dcae} shows a toy structure of three-layer NCAE, which has two hidden layers and one output layer. Formal details are as follows.


\mypar{Input Dropout Module.}Like denoising autoencoder (DAE) \cite{vincent2008extracting}, NCAE learns item correlation patterns via training on a corrupted preference set $\uti$. The only difference is that NCAE only corrupts observed ratings, i.e., non-zero elements of $\ui$. As shown in Fig. \ref{fig:dcae} (left panel), for sparse vector $\uu_4$, only $v_1$, $v_3$, $v_5$ are considered and $v_5$ is finally dropped. This promotes more robust feature learning, since it forces the network to balance the rating prediction and the rating reconstruction. 

In general, corruption strategies include two types: the additive Gaussian noise and the multiplicative drop-out/mask-out noise. In our work, dropout \cite{srivastava2014dropout} is used to distinguish the effects of observed items. For every epoch, the corrupted input $\uti$ is generated from a conditional distribution $p(\uti|\ui)$ and non-zero values in $\ui$ are randomly dropped out (set to 0) independently with probability $q$:
\begin{equation}
\begin{split}
&P(\ut_{id} = \delta \uu_{id}) = 1-q\\
&P(\ut_{id}=0)=q.\\
\end{split}
\end{equation}

To make the corruption unbiased, uncorrupted values in $\ui$ are set to $\delta=1/(1-q)$ times their original values. We find $q=0.5$ usually leads to good results. When predicting the dense $\uhi$, NCAE takes uncorrupted $\ui$ as input.

\mypar{Sparse Forward Module.} In collaborative filtering, the rating matrix $\R$ is usually very sparse, i.e., more than 95\% unobserved values. To handle the sparse input $\ui$ or corrupted $\uti$, NCAE will ignore the effects of unobserved values, in both forward stage and backward stage. For missing values in $\ui$, the input edges (a subset of $W^1$) and the output edges (a subset of $W^L$, when training) will be inhibited by zero-masking.

Formally, in the input layer ($l=0$) and the output layer ($l=L$), there are $N$ nodes for all items. For hidden layer $l<L$, there are $K_l$ nodes, $K_l \ll N$. As shown in Fig. \ref{fig:dcae} (left panel), we only forward $\ut_{41}$, $\ut_{43}$, $\ut_{45}$ with their linking weights. Specifically, when NCAE first maps the input $\uti$ to latent representation $\act^1_i$, we only need to forward non-zero elements of $\uti$ with the corresponding columns of $\W^1$:
\begin{equation}
	\act^1_i = \phi^1(\W^1\uti+ \bias^1)=\phi^1(\sum_{j \in \Ri} \W^1_{*,j}\ut_{ij}),
\end{equation}
where $\W^1_{*,j}$ represents the $j$-th column vector of $W^1$. This reduces the time complexity from $O(K_1N)$ to $O(K_1|\Ri|)$, $|\Ri| \ll N$. For every middle layer, forward and backward procedures follow the same training paradigm like MLP with $O(K_lK_{l-1})$. In the output layer, the latent low-rank representation $\act^{L-1}_i$ is then decoded back to reconstruct the sparse $\ui$ in training or to predict the dense $\uhi$ during inference. Finally, the recommendation list for user $i$ is formed by ranking the unrated items in $\uhi$.  

\mypar{Sparse Backward Module.}To ensure that unobserved ratings do not bring information to NCAE, errors for them are turned to be zero before back-propagation. In other words, no error is back-propagated for missing values via zero-masking\footnote{Otherwise, NCAE learns to predict all ratings as 0, since missing values are converted to zero and the rating matrix $\R$ is quite sparse.}. As shown in Fig. \ref{fig:dcae} (middle panel), unobserved  $v_2$ and $v_4$ are ignored, and error for $v_3$ is computed as $|nn(\uti)_3-{\uu}_{i3}|$, similarly for $v_1$, $v_5$. Furthermore, we can emphasize the prediction criterion and the reconstruction criterion for errors of observed ratings via two hyperparameters $\alpha$ and $\beta$, respectively. Then we employ a new loss function for sparse rating inputs, which separates the two criteria and disregard the loss of unobserved ratings:

\begin{equation}
\label{objective}
\begin{aligned}
	\loss_{2,\alpha,\beta}(\ui,&\uti) =  \frac{\alpha}{|\mathcal{K}(\ui)|} \left( \sum_{j \in \mathcal{C}(\uti)\cap\mathcal{K}(\ui)} [nn(\uti)_{j} - \uu_{ij}]^{2}\right) + \\& \frac{\beta}{|\mathcal{K}(\ui)|} \left( \sum_{j \in \mathcal{K}(\ui)-\mathcal{C}(\uti)} [nn(\uti)_j - \uu_{ij}]^{2}\right),
\end{aligned}
\end{equation}
where $\mathcal{K}(\ui)$ are the indices for observed values of $\ui$, $\mathcal{C}(\uti)$ are the indices for dropped elements of $\uti$. Take the user vector $\uu_4$ for example,  $\mathcal{K}(\uu_4)=\{1,3,5\}$ and $\mathcal{C}(\uu_4)=\{5\}$. Different from \cite{strub2016hybrid}, we divide the loss of DAE by $|\mathcal{K}(\ui)|$ (equals to $|\Ri|$) to balance the impact of each user $i$ on the whole. Actually, NCAE performs better than other methods even when we directly set $\alpha=\beta=1$ for all experiments. Finally, we learn the parameters of NCAE by minimizing the following average loss over all users:
\begin{equation}
 \mathscr{L} = \frac{1}{M} \sum_{i=1}^{M} \loss_{2,\alpha,\beta}(\ui,\uti) + \frac{\lambda}{2} ||\W^+||^2,
\label{eq:loss}
\end{equation}
where we use the squared L2 norm as the regularization term to control the model complexity.

During back-propagation, only weights that are connected with observed ratings are updated, which is common in MF and RBM methods. In practice, we apply Stochastic Gradient Descent (SGD) to learn the parameters. For user $i$, take $\frac{\partial \mathscr{L}}{\partial \W^L_j}$, $\frac{\partial \mathscr{L}}{\partial \bias^L_j}$ and $\frac{\partial \mathscr{L}}{\partial \act^L_i}$ as a example:
\begin{equation}
\frac{\partial \mathscr{L}}{\partial \W^L_j}=\frac{\partial \loss}{\partial \uh_{ij}} \frac{\partial \uh_{ij}}{\partial \W^L_j}+\lambda \W^L_j,
\end{equation}
\begin{equation}
\frac{\partial \mathscr{L}}{\partial \bias^L_j}=\frac{\partial \loss}{\partial \uh_{ij}} \frac{\partial \uh_{ij}}{\partial \bias^L_j}+\lambda \bias^L_j,
\end{equation}
\begin{equation}
\frac{\partial \mathscr{L}}{\partial \act^L_i}= \sum_{j \in \Ri}\frac{\partial \loss}{\partial \uh_{ij}} \frac{\partial \uh_{ij}}{\partial \act^L_i}.
\end{equation}

This leads to a scalable algorithm, where one iteration for user $i$ runs on the order of the number of non-zero entries. The time complexity is reduced from $O(NK_{L-1})$ to $O(|\Ri|K_{L-1})$, $|\Ri| \ll N$.

\mypar{Error Reweighting Module.}As mentioned before, neural network models are easier to overfit on the implicit setting, since only the observed interactions are provided and the model may learn to predict all ratings as 1. To combat overfitting, we introduce a set of variables $c_{ij}$ to determine \textit{which unobserved items a user do not like} and then propose an error reweighting module for implicit NCAE. Specifically, unobserved values of $\ui$ are turned to zero, and errors for them are first computed as $|nn(\uti)_j-0|$, then reweighted by the confidence level $c_{ij}$. In general, we can parametrize $c_{ij}$ based on user exposure \cite{Liangexpo2016} or item popularity \cite{he2016fast}. 

In our work, we implement a simple error reweighting module based on item popularity, since we do not use any auxiliary information (beyond interactions). Similar to \cite{he2016fast}, we assume that popular items not interacted by the user are more likely to be true negative ones. Therefore, unobserved values of $\ui$ on popular items means that the user $i$ is not interested with those items \cite{he2016fast}. We can reweight errors of unobserved places by item popularity $c_j$:
\begin{equation}
	c_j= c_0 \frac{f_j^\omega}{\textstyle \sum_{k=1}^{N}f_k^\omega},
\label{eq:popular}
\end{equation}
where $f_j$ denotes the frequency of item $j$: $|\Rj|/ \sum_{k=1}^N |\mathcal{R}_k|$. $c_0$ is the initial confidence for unobserved values and $\omega$ controls the importance level of popular items. As shown in Fig. \ref{fig:dcae} (right panel), we assume that the item popularity vector is precomputed as [0.17, 0.25, 0.33, 0.17, 0.08], and errors for missing places $v_2$ and $v_4$ are reweighted by 0.25 and 0.17, respectively. Finally, we propose a new loss function for implicit NCAE:
\begin{equation}
\begin{aligned}
\loss_{2}(\ui,\uti) = \sum_{j \in \mathcal{K}(\ui)} [nn(\uti)_{j} \!-\! \uu_{ij}]^{2}\!+\!\!\sum_{j \in \mathcal{J}-\mathcal{K}(\ui)}\!\!c_j[nn(\uti)_j]^2\!,
\end{aligned}
\end{equation}
where $\mathcal{J}=\{1,2,...,N\}$. We set $\alpha=1$ and $\beta=1$, and do not divide the loss by $|\mathcal{K}(x)|$, since it is whole-based optimization for all items of $\ui$. During back-propagation, the time complexity of one user is $O(NK_{L-1})$, which is impractical when the number of items is large. An alternative solution is to build a feature-wise update scheme on the output layer \cite{he2016fast,devooght2015dynamic,yu2012scalable}, which will reduce the complexity to $O((K_{L-1}+|\Ri|)K_{L-1})$.

\subsection{Sparsity-aware Data Augmentation}
In general, the input dropout module and the error reweighting module proposed in Sec. \ref{sec:dca} can provide more item correlation patterns via reweighting the input and the output error, respectively. It is empirically observed that these modules can effectively combat overfitting on the implicit setting. Following this way of thinking, data augmentation strategy can be utilized to provide more item correlation patterns directly with data pre-processing. However, unlike the augmentation on images, which enlarges the dataset using label-preserving transformations like image rotation \cite{simard2003best,krizhevsky2012imagenet}, how to select item correlation patterns \textit{without losing the item ranking order of user preferences} is a key challenge on the implicit setting, since observed interactions are all represented as 1. In general, we have several alternatives to address this problem. For instance, for a particular user, we can reserve items that are viewed the most by him/her or that have similar contents.
\label{sec:sparsity-aware}
\begin{figure}[t]
\centering
\includegraphics[width=8cm]{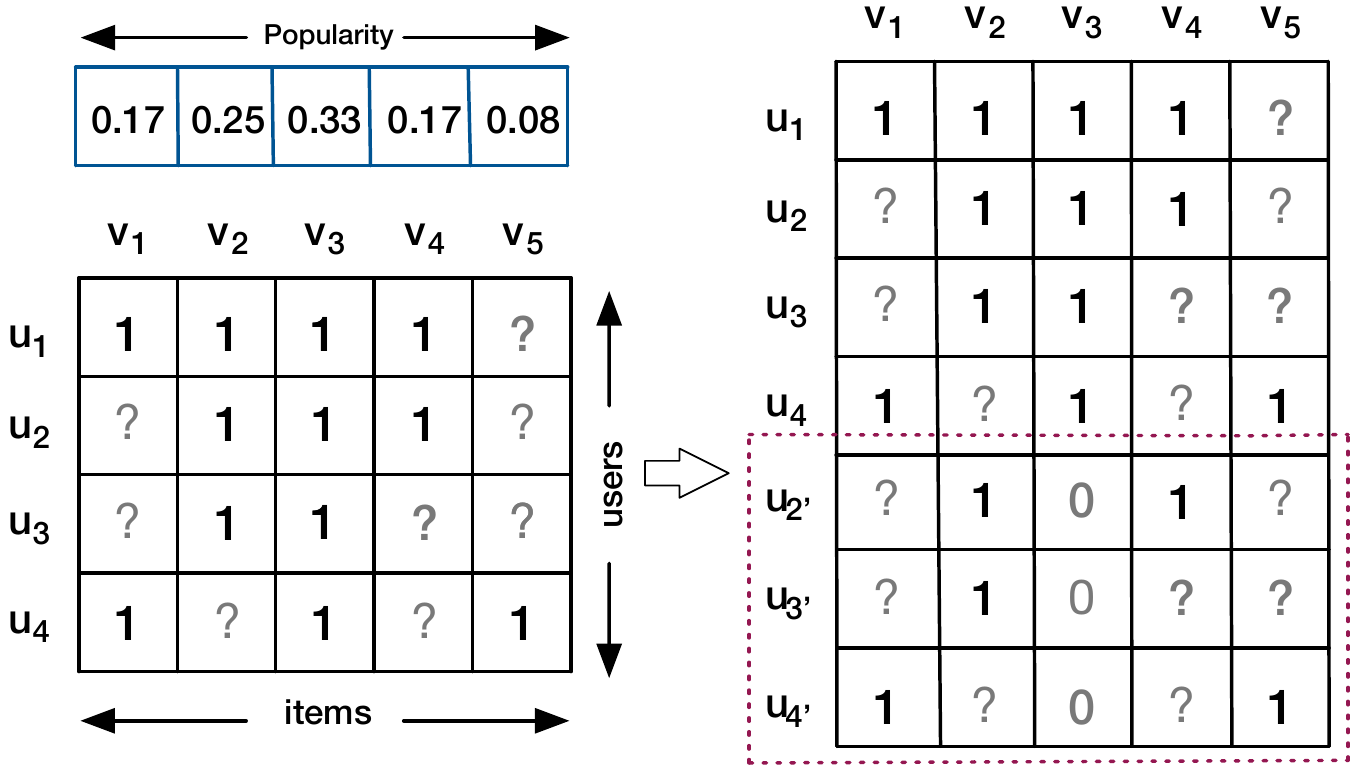}
\vspace{3pt}
\caption{Sparsity-aware data-augmentation: $\uu_2$, $\uu_3$ and $\uu_4$ are augmented when $\epsilon=0.8$ and $p=0.5$. The item popularity vector is pre-computed via Eq. (\ref{eq:popular}) when $\omega=1$ and $c_0=1$.}
\label{fig:data_aug}
\end{figure} 
\begin{figure*}
\centering
\includegraphics[width=18cm]{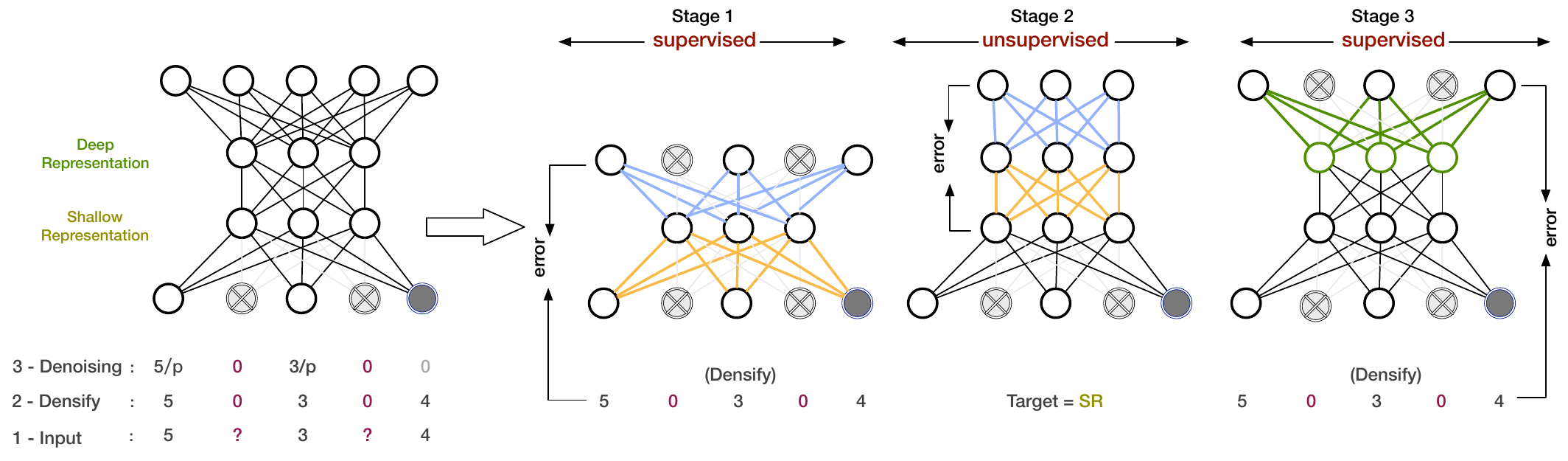}
\vspace{-3pt}
\caption{Left: A three-layer user-based Neural Collaborative Autoencoder with explicit feedback. Right: Proposed greedy layer-wise training mechanism consists of three-stage learning.	In the first two stages, an encoder-decoder paradigm	is performed to learn \textit{SR} and \textit{DR} with corresponding weights. (The encoder part is plotted in orange, the decoder is plotted in blue.) In stage 3, item representation layer (\textit{V}, in green) is added on top of \textit{DR} to perform supervised training.
}
\label{fig:greedy}
\end{figure*}

As mentioned before, we only use the interaction data. Therefore, we employ a distinct form of augmentation based on item popularity, called sparsity-aware data-augmentation. The main assumptions include two points: only the users whose ratings are sparse need to be augmented, which reduces the biases caused by denser users that have interacted with more items; \textit{less popular items interacted by a user are more likely to reveal his/her interest}. Hence, we can drop popular items from the sparse vector $\ui$ to construct augmented sample $\mathbf{u}_{i'}$ and reserve the most relevant item correlation patterns. 

Fig. \ref{fig:data_aug} shows a toy example, where $r_{23}$ is dropped from $u_2$ to construct $u_{2'}$, since $v_3$ is the most popular one that contributes least on the user profile. Specifically, we introduce a sparsity threshold $\epsilon$ and a drop ratio $p$ to control the scale of augmented set. The augmentation procedure follows: If $|\Ri|/N < \epsilon$, we will augment $\ui$ and drop the top $\lfloor|\Ri|*p\rfloor$ popular items to form $\mathbf{u}_{i'}$.

\subsection{Greedy Layer-wise Training}
For explicit feedback, we can employ the NCAE architecture with stacks of autoencoders to learn deep representations. However, learning is difficult in multi-layer architectures. Generally, neural networks can be pre-trained using RBMs, autoencoders or Deep Boltzmann Machines to initialize the model weights to sensible values \cite{vincent2010stacked,hinton2006reducing,salakhutdinov2009deep,le2008representational,bengio2013representation}. Hinton et al. \cite{hinton2006reducing} first proposed a greedy, layer-by-layer unsupervised pre-training algorithm to perform hierarchical feature learning. The central idea is to train one layer at a time with unsupervised feature learning and taking the features produced at that level as input for the next level. The layer-wise procedure can also be applied in a purely supervised setting, called the greedy layer-wise supervised pre-training \cite{bengio2007greedy}, which optimizes every hidden layer with target labels. However, results of supervised pre-training reported in \cite{bengio2007greedy} were not as good as unsupervised one. Nonetheless, we observe that supervised pre-training combined with unsupervised deep feature learning gives the best performance. We now explain how to build a new layer-wise pre-training mechanism for recommender systems, and define:

\mypar{\textit{Definition 1:}}
\textit{Shallow Representation(SR)} for user $i$ is the first layer hidden activation $\act^1_i$ with supervised pre-training, where $\act^1_i=\phi^1(\W^1\uti+ \bias^1)$.

\mypar{\textit{Definition 2:}}
\textit{Deep Representation(DR)} for user $i$ is the $L-1$ layer hidden activation $\act^{L-1}_i$ with unsupervised pre-training using \textit{SR}, where $\act^{L-1}_i=\phi^{L-1}(\W^{L-1}(..\phi^2(\W^2\act^1_i+ \bias^2)..)+\bias^{L-1})$.

\mypar{\textit{Definition 3:}}
\textit{Item Representation(V)} for user-based NCAE is the $L$ layer weight matrix, where $V=\W^L$. Similar definition for \textit{U} of item-based one.

\vspace{0.05in}
Following the definitions, the proposed layer-wise pre-training can be divided into three stages: supervised \textit{SR} learning, unsupervised \textit{DR} learning and supervised \textit{V} learning. Here we consider a three-layer NCAE. Formal details are as follows:

\mypar{Supervised \textit{SR} Learning.}
To train the weights between the input layer and the \textit{SR} layer, we add a decoder layer on top of \textit{SR} to reconstruct the input, as shown in Fig. \ref{fig:greedy}, stage 1. Formally:
\begin{equation}
\begin{split}
	&\act^1_i=\phi^1(\W^1\uti+ \bias^1) \\
	&\act^{'1}_i=\phi^{'1}(\W^{'1}\act^1_i+ \bias^{'1}),\\
\end{split}
\end{equation}
where $\W^{'1}$, $\bias^{'1}$, $\act^{'1}_i$ and $\phi^{'1}$ are the weight matrix, the bias vector, the activation output and the activation function for the first decoder. Similar notations for the decoder of layer $l$ are used in the rest of paper. We use $\act^{'1}_i\!*\!2.25\!+\!2.75$ as estimates ${\uhi}^{'}$ on the explicit setting.

Different from conventional unsupervised pre-training strategies that reconstruct the dense vector of $\uti$ (or reconstruct observed places of $\uti$ without introducing noise), we use the supervised signal $\uu_i$ to form an encoder-decoder scheme. Therefore, we minimize the loss function $\sum_{i=1}^{M}\loss_{2,\alpha,\beta}(\act^{'1}_i,\uu_i)/M$ like Eq. (\ref{eq:loss}) to train $\W^{1}$, $\W^{'1}$, $\bias^{1}$ and $\bias^{'1}$. Supervised \textit{SR} pre-training balances the prediction and the reconstruction criteria in the first layer and encourages the network to discover better item correlation patterns of the sparse input.

\mypar{Unsupervised \textit{DR} Learning.}
After the \textit{SR} is learned, the weights $\W^{1}$ and $\bias^{1}$ of the first layer are frozen, and the decoder part of \textit{SR} is removed. Then we add a new hidden layer with its decoder part on top of \textit{SR} layer to perform unsupervised \textit{DR} learning, as shown in Fig. \ref{fig:greedy}, stage 2. Similar to the unsupervised training procedure proposed by \cite{hinton2006reducing,vincent2010stacked}, only the new added top layer is trained. Thus, we can minimize the square loss between \textit{SR} and corresponding decoder output $\act^{'2}$ to train $\W^{2}$, $\W^{'2}$, $\bias^{2}$ and $\bias^{'2}$:
\begin{equation}
  \mathop{\arg\min}_{\W^{2}, \W^{'2}, \bias^{2},\bias^{'2}} 
  \frac{1}{MK_1} \sum_{i=1}^{M}\sum_{k=1}^{K_1} (\act^{'2}_{ik}-\act^{1}_{ik})^2,
\label{eq:loss_pre2}
\end{equation}
where $K_1$ is the hidden dimension of the \textit{SR} layer. We take average over elements of $\act^{1}_{i}$ to ensure \textit{DR} $\act^{2}_{i}$ can effectively reconstruct each hidden factor of \textit{SR}. Furthermore, \textit{DR} learning can be implemented by multi-layer autoencoders with unsupervised pre-training.

\mypar{Supervised $\mathbf{V}$ Learning.}
In this stage, we perform a supervised pre-training for the top layer (see Fig. \ref{fig:greedy}, green connection of \textit{V}).  The main purpose of this stage is to find a better weight initializer for \textit{V} to reduce the shrinking between layers caused by back-propagating large top error signals when fine-tuning the whole network. Specifically, after \textit{DR} is learned, we fix weights $\W^{2}$ and $\bias^{2}$ along with $\W^{1}$, $\bias^{1}$ (or formally $\W^{L-1}$, $\bias^{L-1}$, ..., $\W^{1}$, $\bias^{1}$), and remove the decoder part. However, there is no need for adding a new decoder part in this stage, for the reason $\act^L_i$ is task-related:
\begin{equation}
	nn(\uti)=\act^L_i=\phi^L(\W^L\act^{L-1}_i+ \bias^L),
\end{equation}
where $L$=3 in this case. Similar to stage 1, supervised signal $\ui$ is used to train \textit{V} and minimize the following average loss function:
\begin{equation}
  \mathop{\arg\min}_{\W^{L}, \bias^{L}} \frac{1}{M} \sum_{i=1}^{M} \loss_{2,\alpha,\beta}(nn(\uti),\uu_i).
\end{equation}

After layer-by-layer pre-training, the parameters of NCAE are usually initialized close to a fairly good solution (see Sec. \ref{sec:experi_explict} for details). Finally, we can perform fine-tuning using supervised back-propagation, minimizing the Eq. (\ref{eq:loss}) corresponding to all parameters $\W^{+}$ to get a better task-related performance. 

\section{Experiments}
\label{sec:experiments}
In this section, we conduct a comprehensive set of experiments that aim to answer five key questions: (1) Does our proposed NCAE outperform the state-of-the-art approaches on both explicit and implicit settings? (2) How does performance of NCAE vary with model configurations? (3) Is the three-stage pre-training useful for improving the expressiveness of NCAE? (4) Do the error reweighting module and the sparsity-aware data augmentation work on the implicit setting? (5) Is NCAE scalable to large datasets and robust to sparse data?
\begin{table}[t]
  \centering
  \setlength{\abovecaptionskip}{3pt}
  \caption{Statistics of datasets}
  \label{tab:data}
  \begin{tabular}{ccccc}
    \toprule
    Dataset&\#users&\#items&\#ratings&sparsity\\
    \midrule
    ML-10M & 69,878 & 10,073 & 9,945,875 & 98.59\%\\
    Delicious & 1,867 & 69,226 & 104,799 & 99.92\%\\
    Lastfm & 1,892 & 17,632 & 92,834 & 99.72\%\\
  	\bottomrule
\end{tabular}
\end{table}

\subsection{Experimental Setup}
We first describe the datasets and the training settings, and then elaborate two evaluation protocols, including evaluation metrics, compared baselines and parameter settings.

\mypar{Datasets.}
To demonstrate the effectiveness of our models on both explicit and implicit settings, we used three real-world datasets obtained from MovieLens 10M\footnote{\url{https://grouplens.org/datasets/movielens/10m}}, Delicious\footnote{\url{https://grouplens.org/datasets/hetrec-2011}} and Lastfm\footnotemark[5]. MovieLens 10M dataset consists of user's explicit ratings on a scale of 0.5-5. For Delicious and Lastfm datasets, we consider a user feedback for an item as ‘1’ if the user has bookmarked (or listened) the item. Table \ref{tab:data} summarizes the statistics of datasets.

\mypar{Training Settings.}
We implemented NCAE using Python and Tensorflow\footnote{\url{https://www.tensorflow.org}}, which will be released publicly upon acceptance. Weights of NCAE are initialized using Xavier-initializer \cite{glorot2010understanding} and trained via SGD with a mini-batch size of 128. Adam optimizer \cite{kingma2014adam} is used to adapt the step size automatically with the learning rate set to 0.001. We use grid search to tune other hyperparameters of NCAE and compared baselines on a separate validation set. Then the models are retrained on the training plus the validation set and finally evaluated on the test set. We repeat this procedure 5 times and report the average performance.

\mypar{Methodology.}
We evaluate with two protocols: 

\textbf{- Explicit protocol.} We randomly split MovieLens 10M dataset into 80\%-10\%-10\% training-validation-test datasets. Since ratings are 10-scale, we use Root Mean Square Error (\textit{RMSE}) to evaluate prediction performance \cite{lee2013local,strub2016hybrid,sedhain2015autorec}:
\begin{equation}
\begin{aligned}
\textit{RMSE}= \sqrt{\frac{1}{|\mathbf{R}_{te}|}\sum_{\tau=1}^{|\mathbf{R}_{te}|}{\Big(R_\tau -\hat {R_\tau} \Big)^2}}
\end{aligned}
\end{equation}
where $|\mathbf{R}_{te}|$ is the number of ratings in the test set, $\R_\tau$ is the real rating of an item and $\hat{R_\tau}$ is its corresponding predicted rating. For the explicit setting, we compare NCAE with the following baselines:
\begin{itemize}
\item \textbf{ALS-WR} \cite{zhou2008large} is a classic MF model via iteratively optimizing one parameter with others fixed.
\item \textbf{BiasedMF} \cite{koren2009matrix} incorporates user/item biases to MF, which performs gradient descent to update parameters.
\item \textbf{SVDFeature} \cite{chen2012svdfeature} is a machine learning toolkit for feature-based collaborative filtering, which won the KDD Cup for two consecutive years.
\item \textbf{LLORMA} \cite{lee2013local} usually performs best among conventional explicit methods, which relaxes the low-rank assumption of matrix approximation.
\item \textbf{I-AutoRec} \cite{lee2013local} is a one-hidden layer neural network, which encodes sparse item preferences and aims to reconstruct them in the decoder layer.
\item \textbf{V-CFN} \cite{strub2016hybrid} is a state-of-the-art model for explicit feedback, which is based on DAE. This model can be considered as an extension of AutoRec.  
\end{itemize}

In the experiments, we train a 3-layer item-based NCAE for MovieLens 10M dataset, which has two hidden layers with equal hidden factors $K$. We employ $K$ as 500 unless otherwise noted. For regularization, we set weight decay $\lambda=0.0002$ and input dropout ratio $q=0.5$. Parameter settings for compared baselines are the same in original papers. We compare NCAE with the best results reported in authors' experiments under the same 90\% (training+validation)/10\% (test) data splitting procedure. 

\vspace{0.5\baselineskip}
\textbf{- Implicit protocol.} Similar to \cite{he2016fast}, we adopt the leave-one-out evaluation on Delicious and Lastfm datasets, where the latest interaction of each user is held out for testing. To determine the hyperparameters, we randomly sample one interaction from the remaining data of each user as the validation set. We use Hit Ratio (\textit{HR}) and Normalized Discounted Cumulative Gain (\textit{NDCG}) for top-M evaluations, and report the score averaged by all users. Without special mention, we set the cut-off point to 100 for both metrics. We compare NCAE with the following top-M methods:
%

\begin{itemize}
\item \textbf{POP} is a non-personalized model that rank items by their popularity. We use the implementation in \cite{rendle2009bpr}.
\item \textbf{BPR} \cite{rendle2009bpr} samples negative interactions and optimizes the MF model with a pairwise ranking loss. We set the learning rate and weight decay $\lambda$ to the best values 0.1 and 0.01.
\item \textbf{WMF} \cite{hu2008collaborative} treats all unobserved interactions as negative instances, weighting them uniformly via a confidence value $c$. We set $c$ and $\lambda$ to the best values 0.05 and 0.1.
\item \textbf{eALS} \cite{he2016fast} is a state-of-the-art MF-based model, weighting all unobserved interactions non-uniformly by item popularity. We set $\lambda$ to the best value 0.05.
\item \textbf{NCF} \cite{he2017neural} is a state-of-the-art neural network model for implicit feedback that utilizes a multi-layer perceptron to learn the interaction function. Similar to \cite{he2017neural}, we employ a 4-layer MLP with the architecture of $2K{\rightarrow}K{\rightarrow}K/2{\rightarrow}1$, where the embedding size for the GMF component is set to $K$. 
\end{itemize}

We train a 2-layer user-based NCAE for the two datasets. Since it is easier to overfit the training set on the implicit setting, we set input dropout ratio $q=0.5$ and weight decay $\lambda=0.01$, and further utilize the error reweighing module and the sparsity-aware data augmentation strategy. Similar to \cite{he2016fast}, we compute the item popularity using empirical parameters ($c_0=512$ and $\omega=0.5$) for both NCAE and eALS. For a fair comparison, we report the performance of all methods when $K$ is set to [8, 16, 32, 64, 128]. We employ $K$ as 128 for NCAE unless otherwise noted.

\subsection{Explicit Protocol}
\label{sec:experi_explict}
We first study how model configurations impact NCAE's performance. Then we demonstrate the effectiveness of pre-training. Finally we compare NCAE with explicit baselines.

\mypar{1) Analysis of Model Configurations}

\begin{figure*}
\centering
\subfigure[Training \textit{RMSE}]{
\begin{minipage}{4cm}
\centering
\includegraphics[scale=0.28]{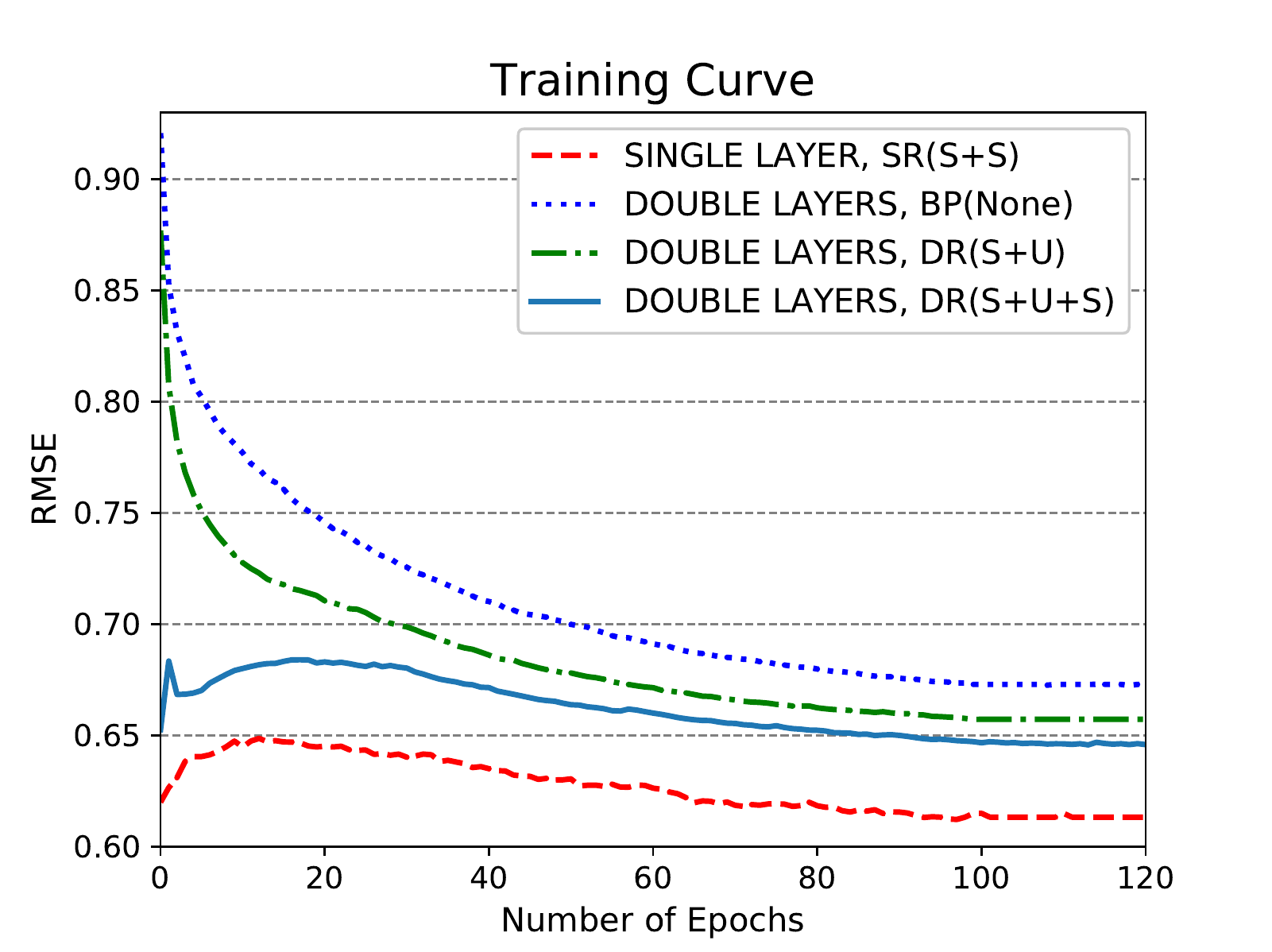}
\end{minipage}
}
\subfigure[Test \textit{RMSE}]{
\begin{minipage}{4cm}
\centering
\includegraphics[scale=0.28]{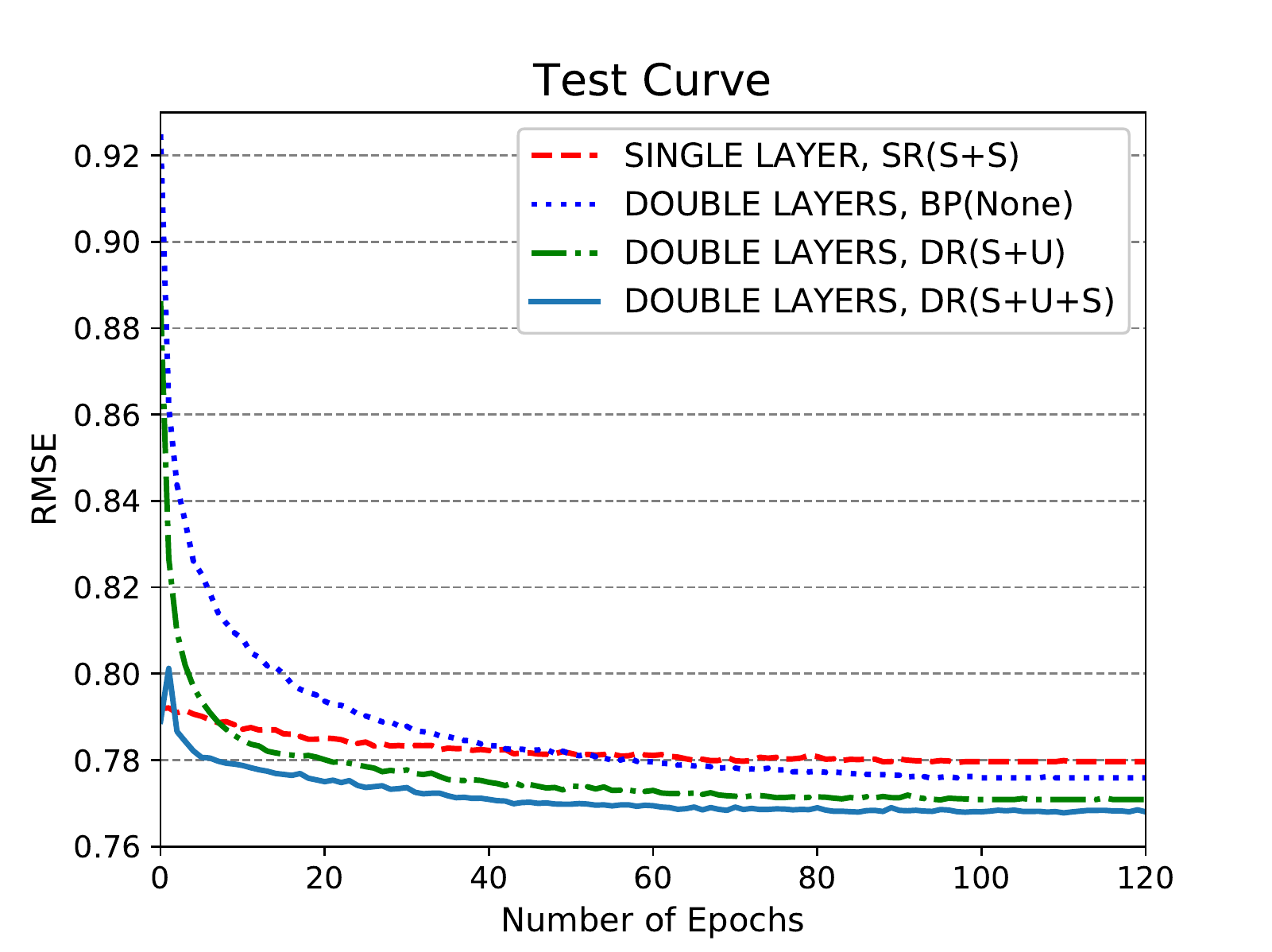}
\end{minipage}
}
\subfigure[Training \textit{RMSE}]{
\begin{minipage}{4cm}
\centering
\includegraphics[scale=0.28]{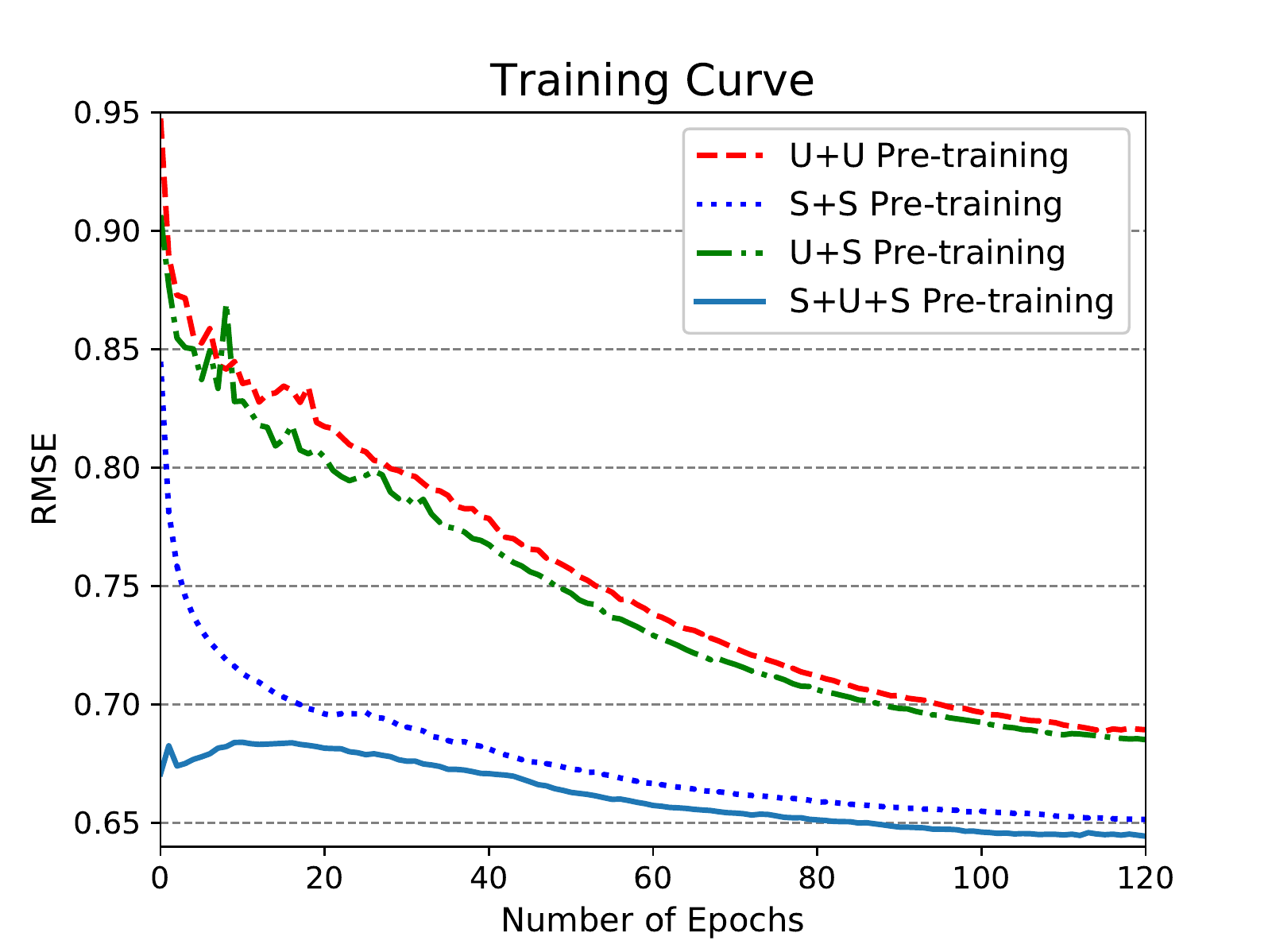}
\end{minipage}
}
\subfigure[Test \textit{RMSE}]{
\begin{minipage}{4cm}
\centering
\includegraphics[scale=0.28]{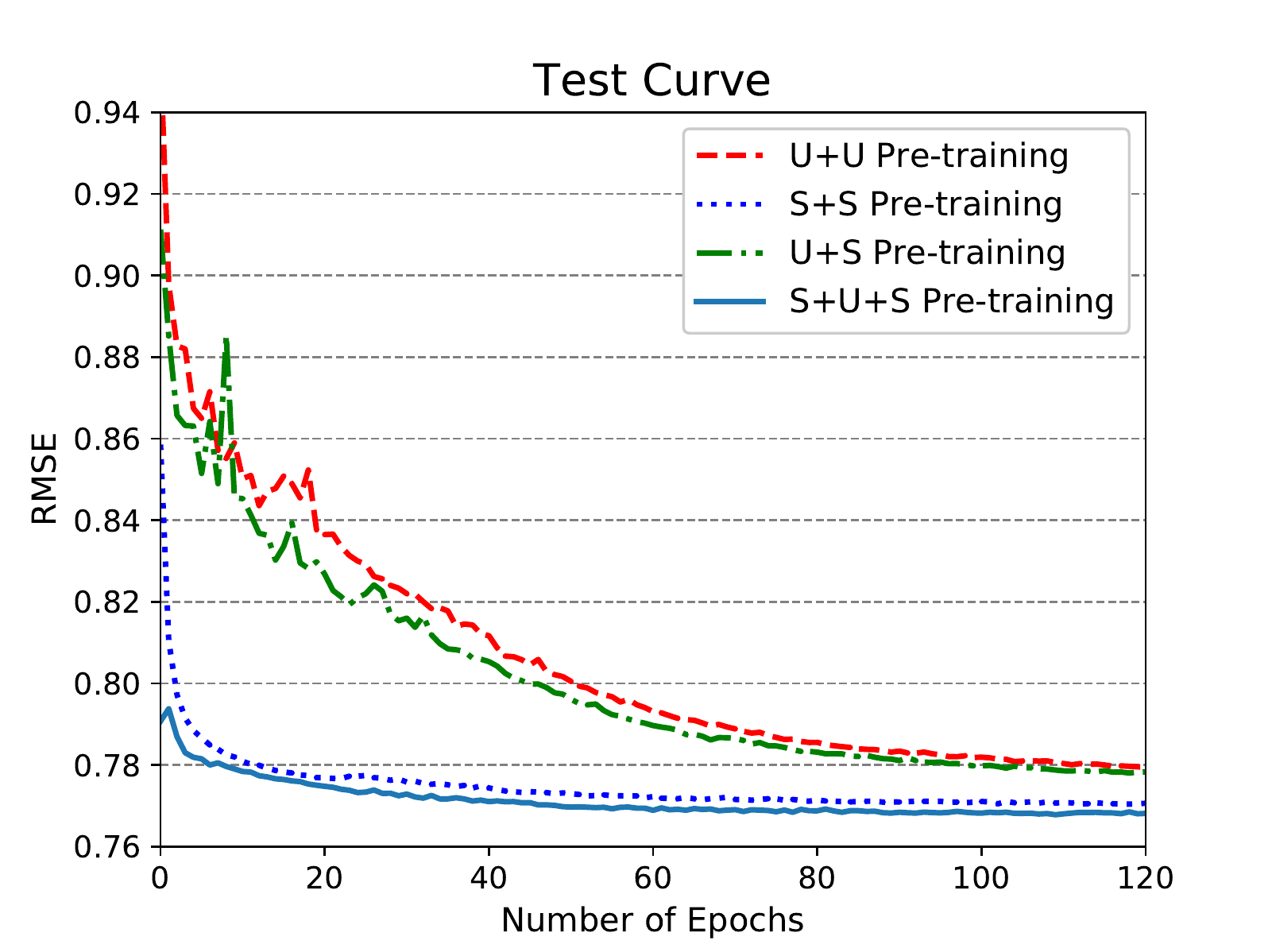}
\end{minipage}
}
\vspace{-10pt}
\caption{Utility of three-stage pre-training on MovieLens 10M. (a, b) Effect of \textit{SR}, \textit{DR} and \textit{V} pre-training components. (c, d) Performance comparison with layer-wise unsupervised pre-training  and layer-wise supervised pre-training strategies.}
\label{fig:experi_pretraining}
\end{figure*}
Table \ref{tab:configurations} shows the performance of NCAE with different configurations, where "SINGLE" means one hidden layer, "*" means the utilization of our pre-training mechanism, and the size of hidden factors $K$ is set to [200, 300, 400, 500]. It can be observed that increasing the size of hidden factors is beneficial, since a larger $K$ generally enlarges the model capacity. Without pre-training, NCAE with two hidden layers may perform worse than that with one hidden layer, which is due to the difficulty on training deep architectures. NCAE with pre-training performs better than that without pre-training, especially when a deeper architecture is used.
\begin{table}[t]
  \centering
  \setlength{\abovecaptionskip}{3pt}
  \setlength{\belowcaptionskip}{0pt}
  \caption{Test \textit{RMSE} with varying configurations}
  \label{tab:configurations}
  \begin{tabular}{lcccc}
    \toprule
    Configurations&200&300&400&500\\
    \midrule
    \quad SINGLE & 0.791 & 0.784 & 0.781 & 0.779\\
    \quad SINGLE* & 0.790 & 0.783 & 0.780 & 0.778\\
   	\quad DOUBLE & 0.803 & 0.788 & 0.781 & 0.775\\
    \quad DOUBLE* & 0.794 & 0.779 & 0.771 & \textbf{0.767}\\
  	\bottomrule
\end{tabular}
\end{table}

\mypar{2) Utility of Pre-training}

To demonstrate the utility of three-stage pre-training, we conducted two sub-experiments: effectiveness analysis for \textit{SR}, \textit{DR} and \textit{V} components; performance comparison with conventional pre-training strategies. Firstly, we term "U" as "unsupervised" and "S" as "supervised". Thus, "S+U+S" means our proposed three-stage pre-training method, and "S+U" means "supervised + unsupervised" pre-training for the first two layers. Besides, "BP(None)" means NCAE without pre-training and "DOUBLE" means NCAE with two-hidden layers (3-layer NCAE). For a fair comparison, we set layer-wise pre-training epochs to 10 for each layer of compared strategies. It takes nearly 6.3 seconds per epoch.

Fig. \ref{fig:experi_pretraining}(a,b) shows the training/test \textit{RMSE} curves of NCAE on different pre-training configurations, where epoch-0 outputs \textit{RMSE} after pre-training. From it, we get: 
\begin{itemize}
\item Compared with "BP(None)", "S+U+S" gets lower \textit{RMSE} on epoch-0 in both learning and inferring phases, and brings better generalization after fine-tuning, indicating  the usefulness of pre-training for initializing model weights to a good local minimum.
\item "S+U+S" gets better generalization on the test set than one-hidden layer NCAE with "S+S", verifying the necessity of \textit{DR} to learn high-level representation of the input. Besides, "S+S" performs best on the training set, but gets worst \textit{RMSE} on the test set (even when we set a larger weight decay $\lambda$), which may be caused by overfitting or its model capacity. 
\item "S+U+S" performs better than "S+U" at the beginning of learning phase, indicating that \textit{V} pre-training can keep the top error signal in a reasonable range, and thus reduce weight shrinking in the lower layers when fine-tuning. Moreover, \textit{RMSE} of "S+U+S" increases slightly at the beginning, which is in accord with the hypothesis of shrinking.
\end{itemize}

Fig. \ref{fig:experi_pretraining}(c,d) shows the performance of 3-layer NCAE with different pre-training strategies. From it, we get: 
\begin{itemize}
\item "U+U" performs worst in both learning and inferring phases. The greedy layer-wise unsupervised pre-training do not emphasize the prediction criterion. Therefore, the network cannot preserve the task-related information as the layer grows (compared to "U+S" and "S+S") due to the sparsity nature of the recommendation problem.
\item "S+U+S" outperforms "S+S" at the beginning of learning and finally gets similar generalization power. This shows the effectiveness of greedy layer-wise supervised pre-training. "S+S" also performs similarly to "S+U" (see DR(S+U)). However, since $M$ and $N$ are usually large, i.e., $M\gg K$ and $N\gg K$, supervised pre-training for all layers is time-consuming and space-consuming. Besides, "S+U+S" can reduce the shrinking of weights.
\item Comparing "U+S" with "S+S" and "S+U+S", we can observe that supervised \textit{SR} pre-training in the first layer is critical to the final performance; \textit{SR} learning balances the prediction criterion and the reconstruction criterion, encouraging the network to learn a better representation of the sparse input.
\end{itemize}

\mypar{3) Comparisons with Baselines}
\begin{table}[t]
  \centering
  \setlength{\abovecaptionskip}{3pt}
  \caption{Test \textit{RMSE} on ML-10M (90\%/10\%).}
  \label{tab:explict_baselines}
  \begin{tabular}{p{4cm} p{1cm}}
    \toprule
    \quad Algorithms&\textit{RMSE}\\
    \midrule
    \quad BiasedMF & 0.803\\
    \quad ALS-WR & 0.795\\
    \quad SVDFeature & 0.791\\
    \quad LLORMA & 0.782\\
    \quad I-AutoRec & 0.782\\
    \quad V-CFN & 0.777\\
    \quad NCAE & 0.775 \\
	\quad NCAE* & \textbf{0.767}\\
  	\bottomrule
\end{tabular}
\end{table}
Table \ref{tab:explict_baselines} shows the performance of NCAE and other baselines on ML-10M, where the results of baselines are taken from original papers under the same 90\%/10\% data splitting. We conducted one-sample paired $t$-tests to verify that  improvements of NCAE and NCAE* are statistically significant for $sig<0.005$. NCAE without pre-training has already outperformed other models, which achieves \textit{RMSE} of 0.775. Note that the strong baseline CFN shares a similar architecture like NCAE, but employs well-tuned parameters ($\alpha$ and $\beta$) and additional input pre-processing. The new training objective of NCAE (Eq. \ref{objective}, which balances the impact of each user) provides further performance gain even without these procedures. The performance can be further enhanced by using three-stage pre-training, with a test \textit{RMSE} of 0.767. This demonstrates that the proposed pre-training method can enhance the expressiveness of NCAE, bringing better generalization.
\subsection{Implicit Protocol}
\label{sec:implicit_protocol}
We first study how the sparsity-aware data augmentation impacts NCAE's performance. Then we compare NCAE with several top-M methods in detail. 
\begin{table}[t]
  \caption{Performance of NCAE and NCAE+ on different clusters of users sorted by their ratings.}
  \label{tab:scale}
  \centering
  \begin{tabular}{l|c|c|c|c}
    \toprule
    &\multicolumn{2}{c|}{ \textbf{NCAE}}&\multicolumn{2}{c}{ \textbf{NCAE+}} \\
	\midrule
    Interval  &\textit{\small{HR@100}} & \textit{\small{NDCG@100}} & \textit{\small{HR@100}} & \textit{\small{NDCG@100}} \\ 
    \midrule
    0-1(55) &0.055& 0.025 &0.109 &0.064\\
    1-5(68) &0.279 &0.072 & 0.338 &0.136\\
    5-10(55) &0.564& 0.145& 0.555& 0.253\\
    10-20(96)& 0.323& 0.076&0.333& 0.154\\
    20-30(78) &0.263& 0.078& 0.282& 0.103\\
    \bottomrule
  \end{tabular}
\end{table}

\begin{figure}[t]
\centering
\subfigure[\textit{NDCG}@100 (effect of $\epsilon$, $p$)]{
\begin{minipage}{4.0cm}
\centering
\includegraphics[scale=0.27]{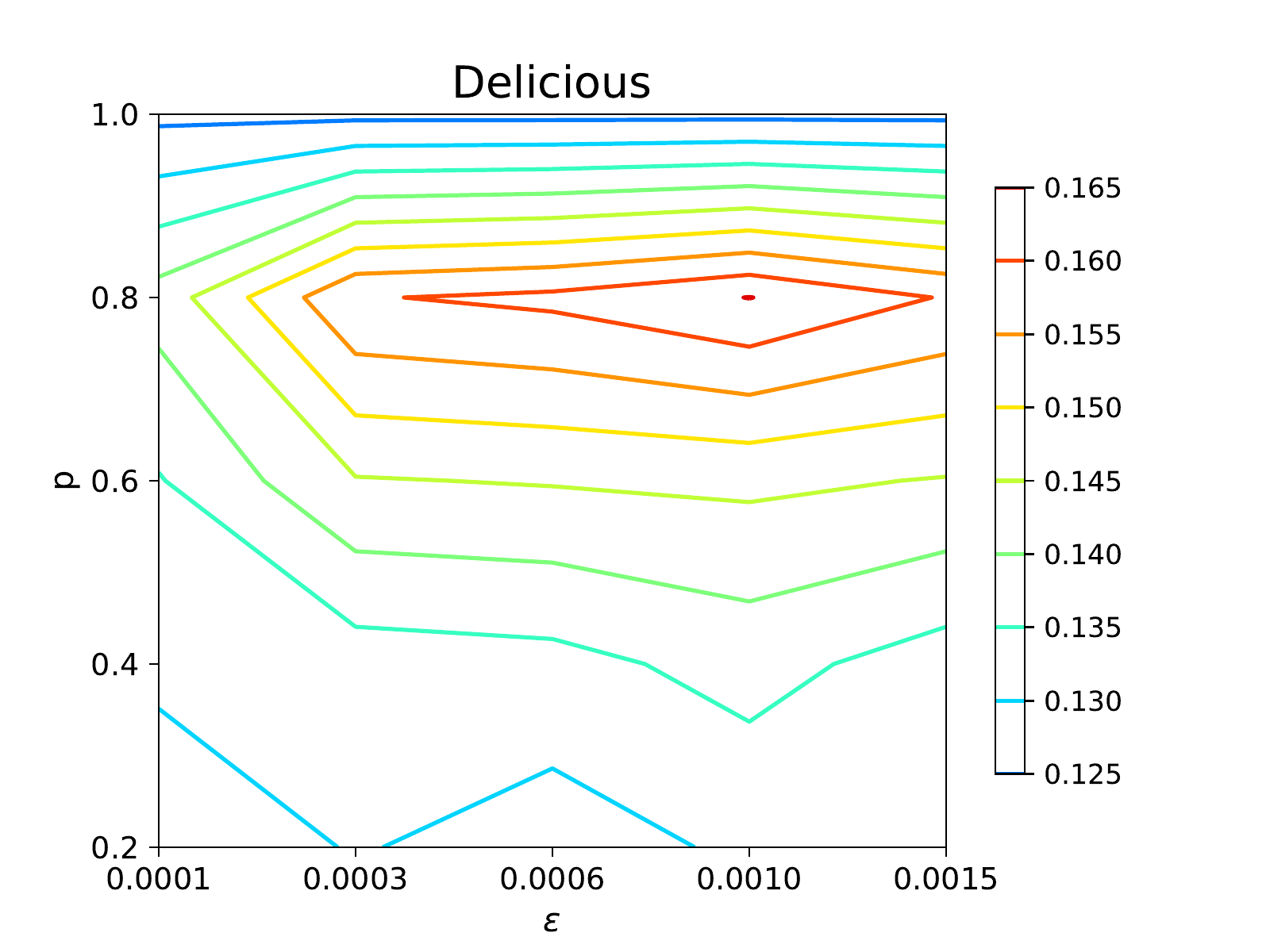}
\end{minipage}
}
\subfigure[\textit{HR}@100 ($\epsilon=0.001$, $p=0.8$)]{
\begin{minipage}{4cm}
\centering
\includegraphics[scale=0.27]{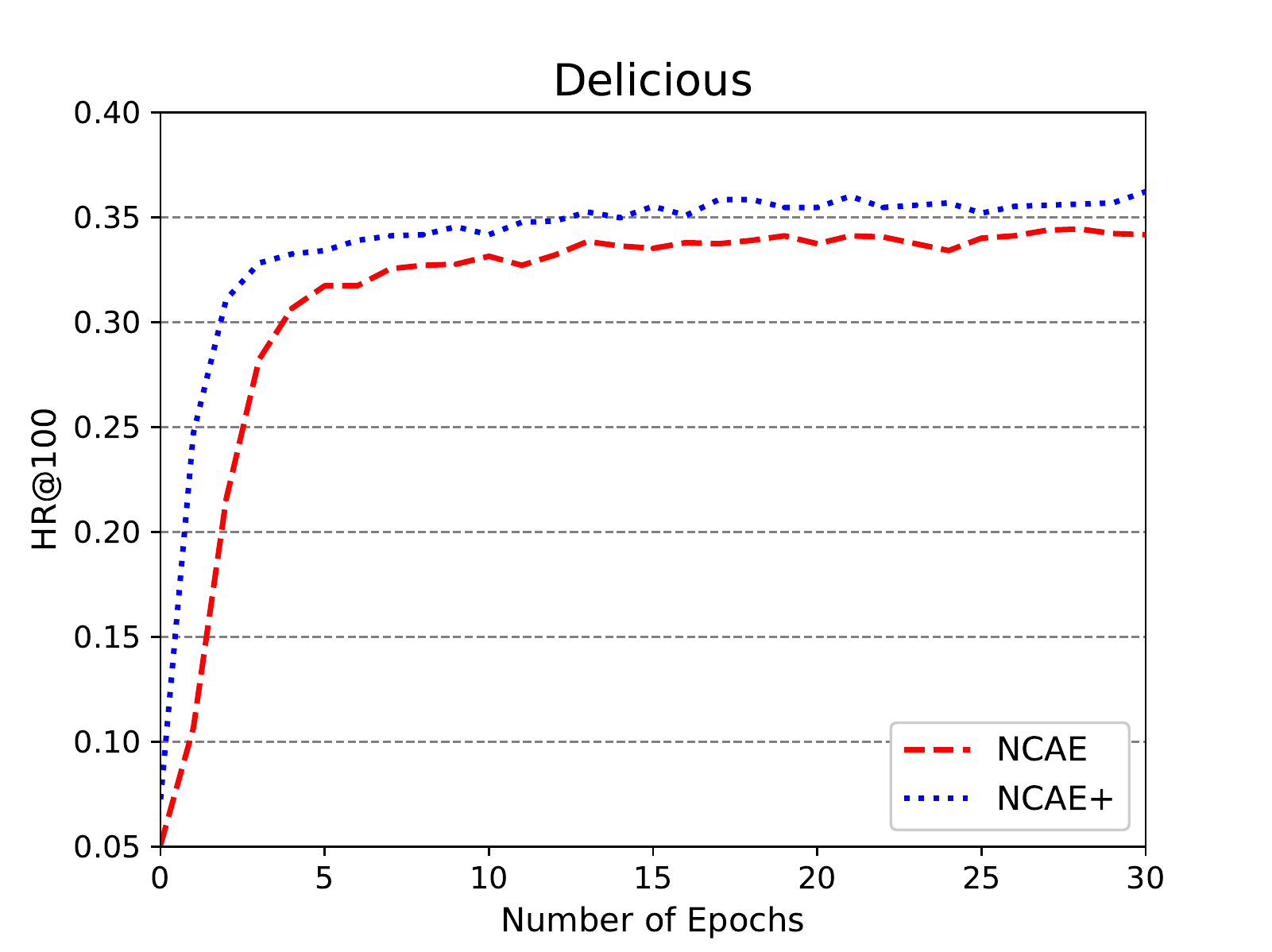}
\end{minipage}
}
\vspace{-6pt}
\caption{Effect of data augmentation on Delicious ($K$=128.)}
\label{fig:experi_data_aug}
\end{figure}

\begin{figure*}
\centering
\subfigure[Delicious --- \textit{HR}@100]{
\begin{minipage}{8cm}
\centering
\includegraphics[width=2.3in]{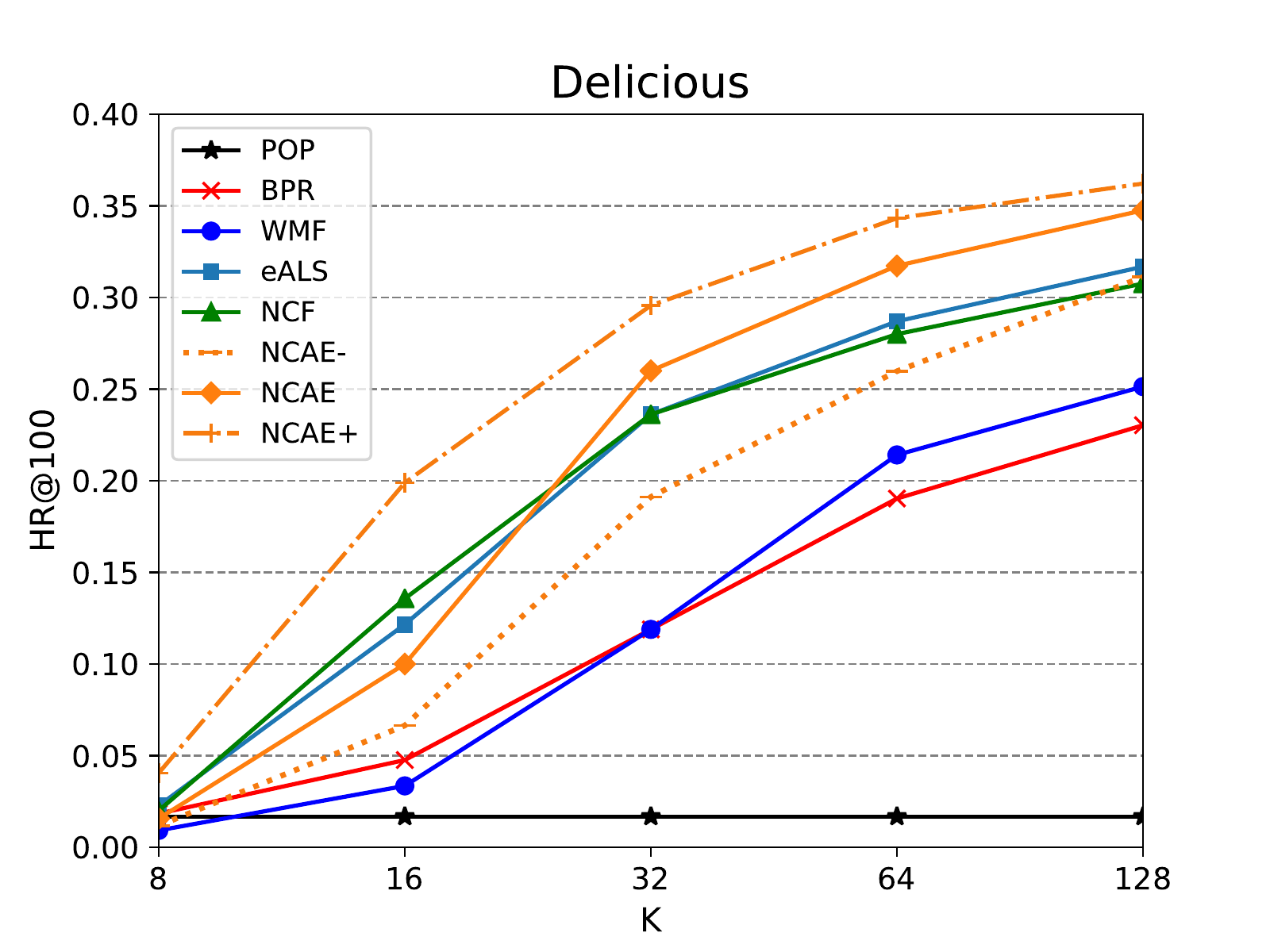}
\end{minipage}
}
\subfigure[Delicious --- \textit{NDCG}@100]{
\begin{minipage}{8cm}
\centering
\includegraphics[width=2.3in]{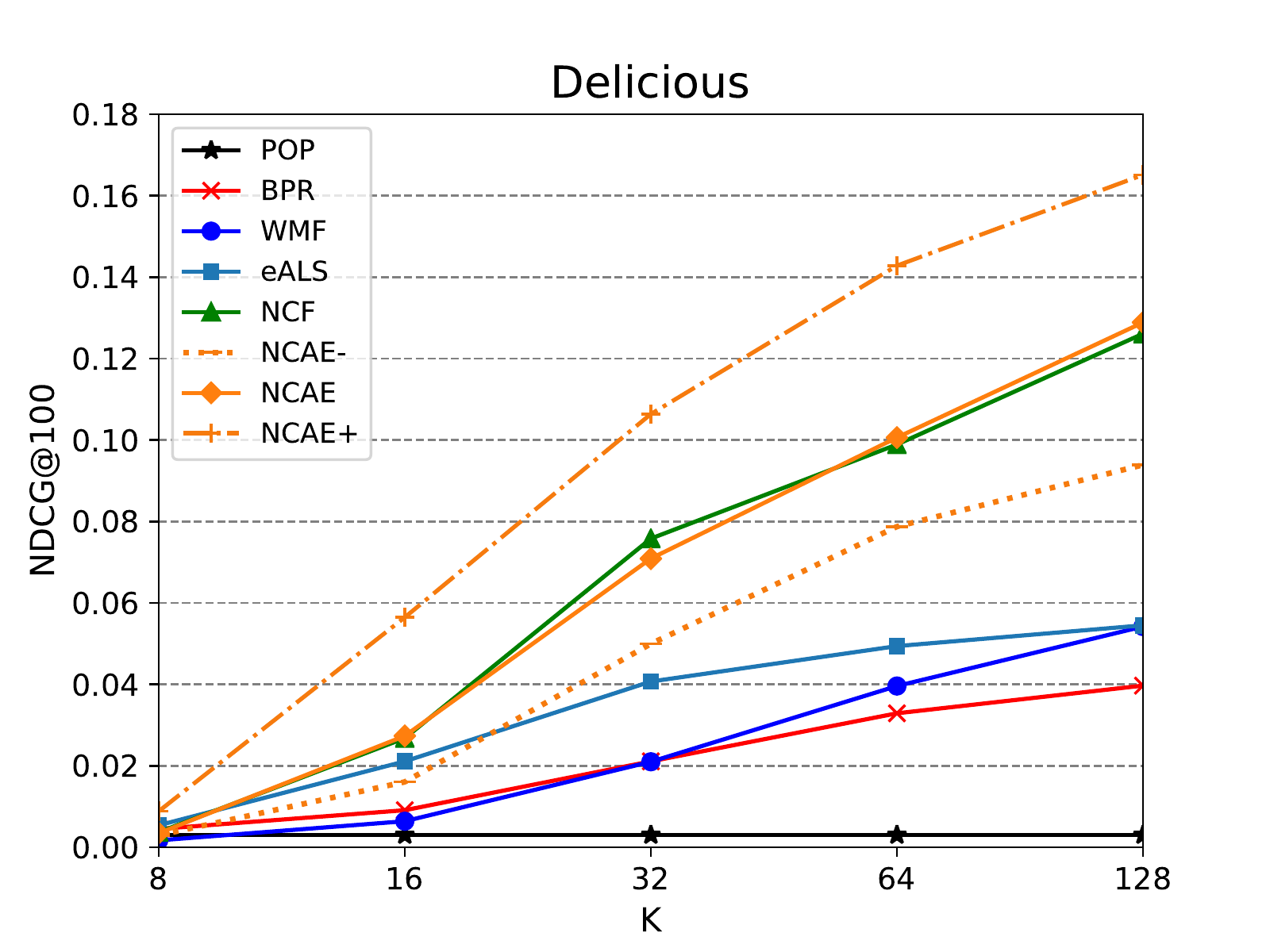}
\end{minipage}
}
\vspace{-3pt}
\caption{Performance of \textit{HR}@100 and \textit{NDCG}@100 w.r.t. the number of hidden factors on Delicious dataset.}
\label{fig:experi_implict_1}
\end{figure*}
\begin{figure*}
\centering
\subfigure[Lastfm --- \textit{HR}@100]{
\begin{minipage}{8cm}
\centering
\includegraphics[width=2.3in]{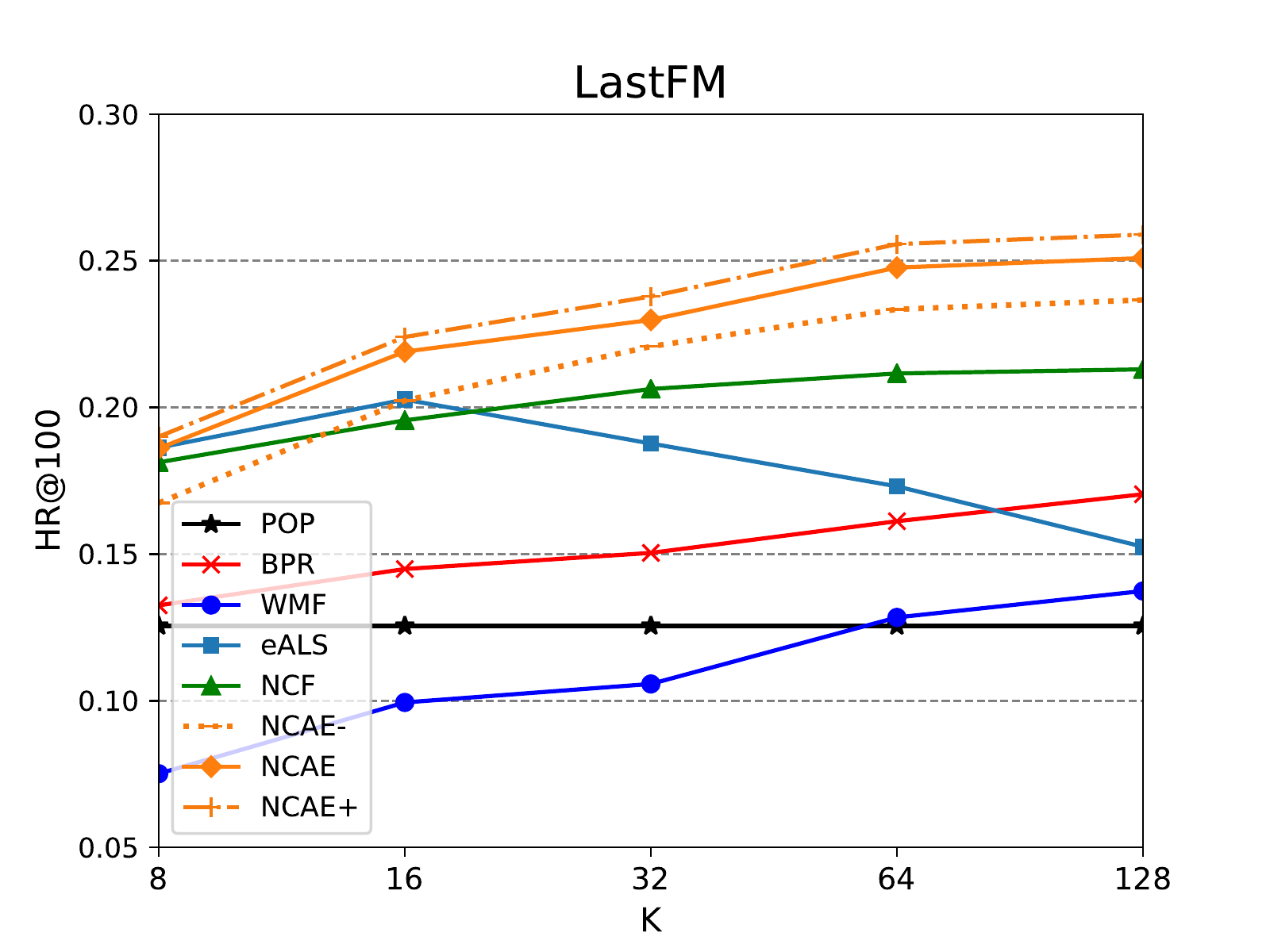}
\end{minipage}
}
\subfigure[Lastfm --- \textit{NDCG}@100]{
\begin{minipage}{8cm}
\centering
\includegraphics[width=2.3in]{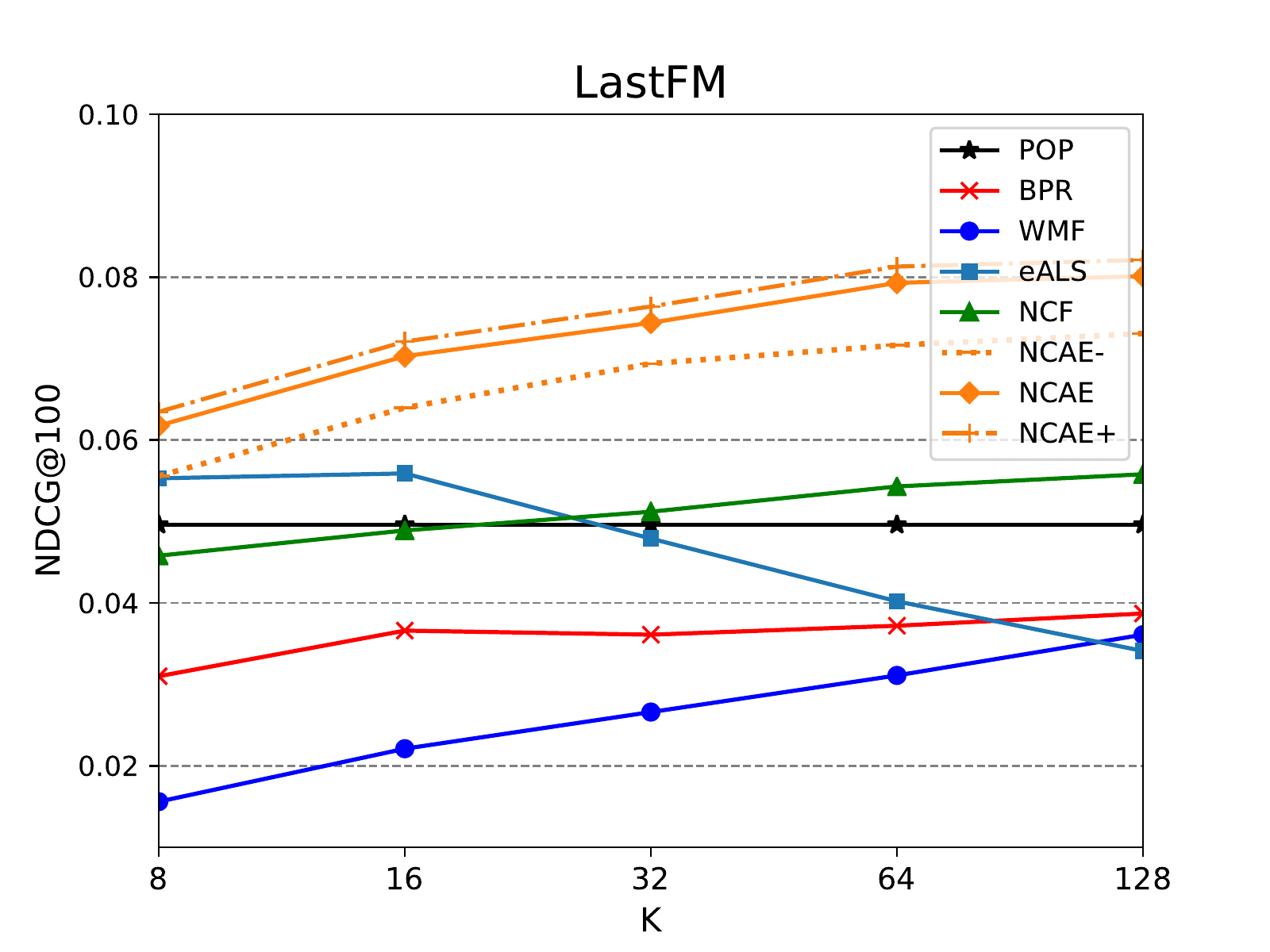}
\end{minipage}
}
\vspace{-3pt}
\caption{Performance of \textit{HR}@100 and \textit{NDCG}@100 w.r.t. the number of hidden factors on Lastfm dataset.}
\label{fig:experi_implict_2}
\end{figure*}

\begin{figure*}
\centering
\subfigure[Delicious --- \textit{HR}@M]{
\begin{minipage}{8cm}
\centering
\includegraphics[width=2.3in]{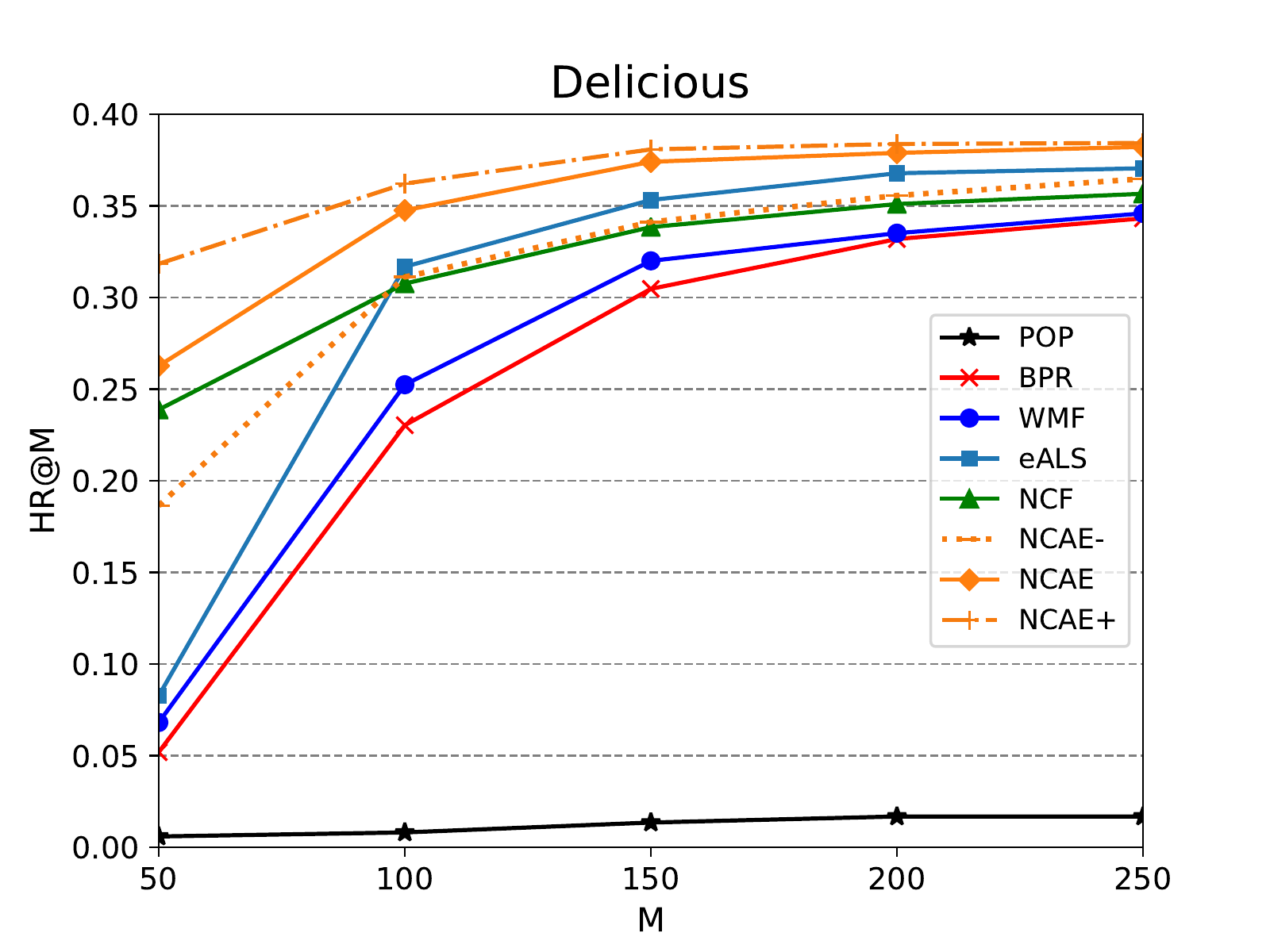}
\end{minipage}
}
\subfigure[Delicious --- \textit{NDCG}@M]{
\begin{minipage}{8cm}
\centering
\includegraphics[width=2.3in]{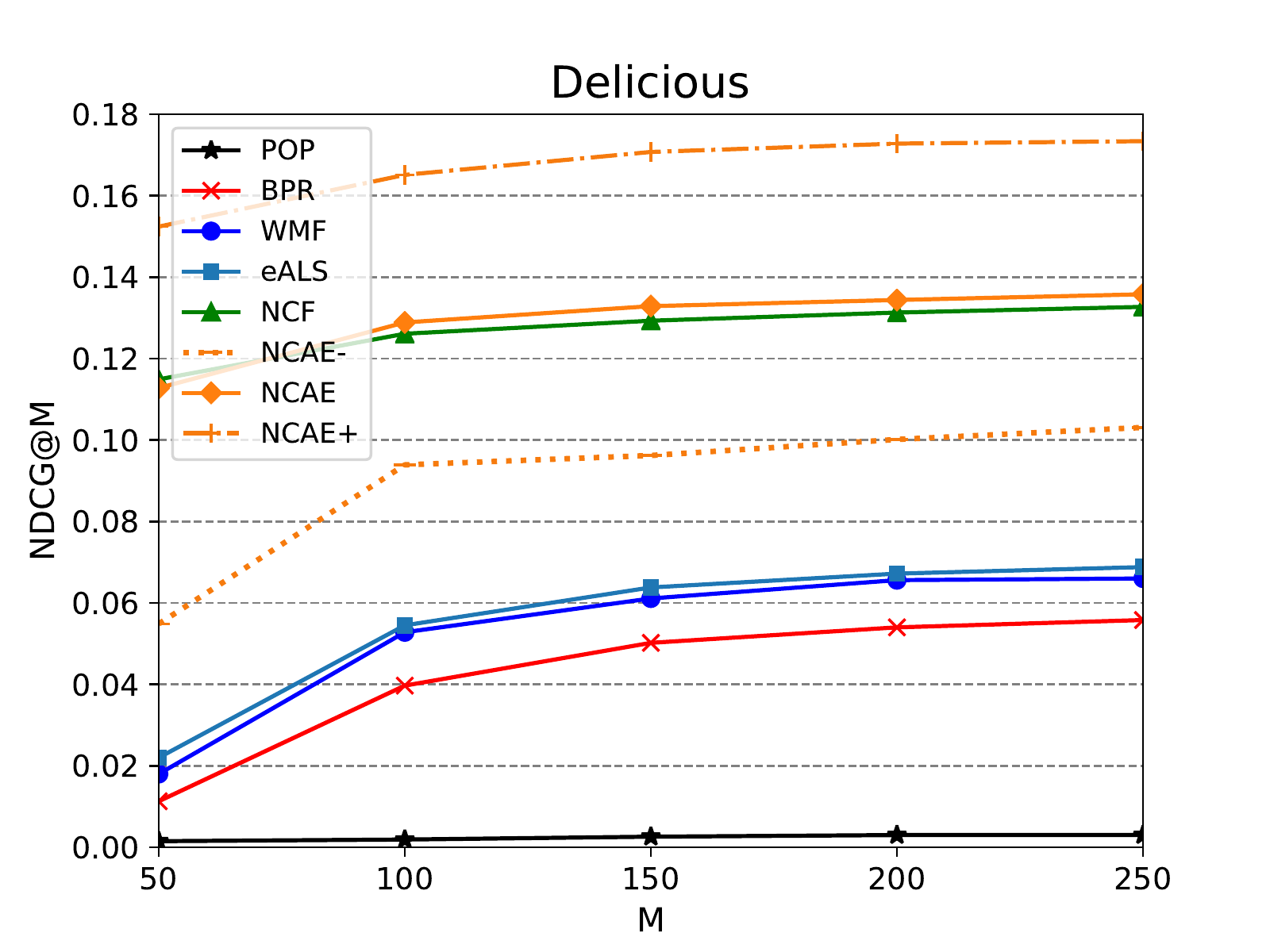}
\end{minipage}
}
\vspace{-3pt}
\caption{Evaluation of Top-M item recommendation where M ranges from 50 to 250 on Delicious dataset.}
\label{fig:experi_implict_3}
\end{figure*}
\begin{figure*}
\centering
\subfigure[Lastfm --- \textit{HR}@M]{
\begin{minipage}{8cm}
\centering
\includegraphics[width=2.3in]{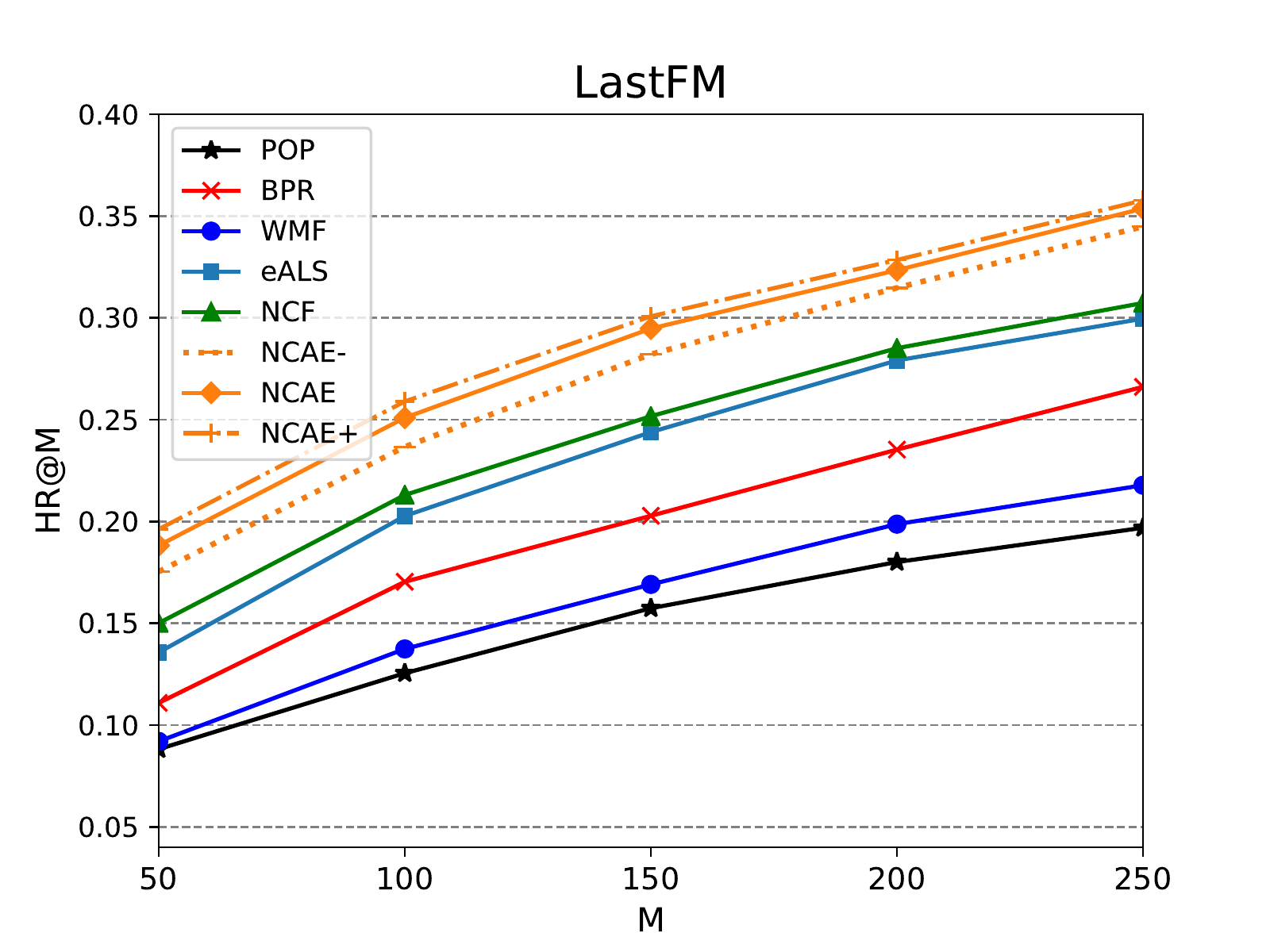}
\end{minipage}
}
\subfigure[Lastfm --- \textit{NDCG}@M]{
\begin{minipage}{8cm}
\centering
\includegraphics[width=2.3in]{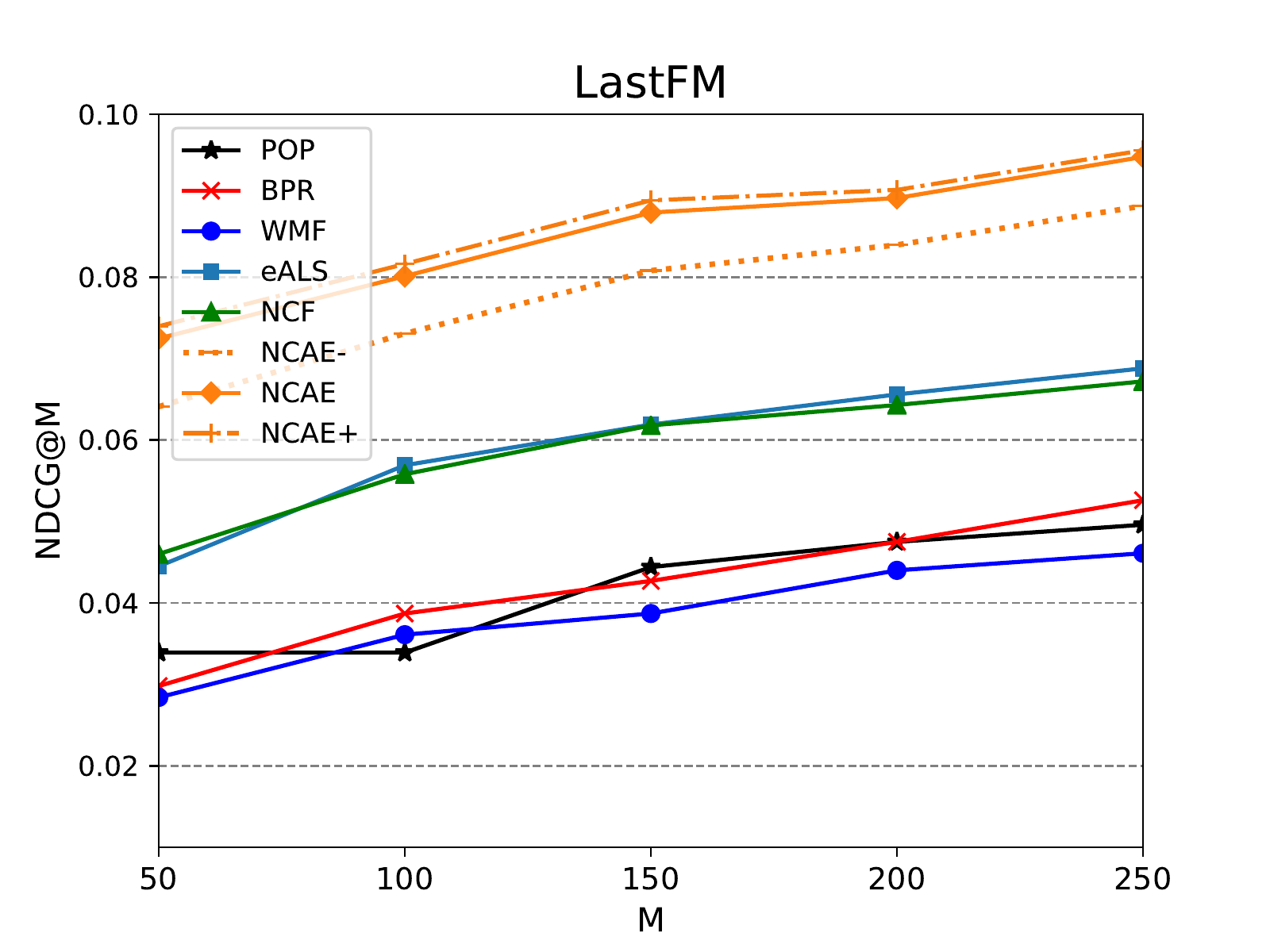}
\end{minipage}
}
\vspace{-3pt}
\caption{Evaluation of Top-M item recommendation where M ranges from 50 to 250 on Lastfm dataset.}
\label{fig:experi_implict_4}
\end{figure*}

To illustrate the impact of our data-augmentation strategy, we show the performance of NCAE w.r.t. sparsity threshold $\epsilon$ and drop ratio $p$ in Fig. \ref{fig:experi_data_aug}(a). Noticeably, $p=1.0$ means that we drop all ratings of one user, i.e., we do not use data-augmentation in this case (with a \textit{NDCG}@100 of 0.125). NCAE achieves the best performance when $\epsilon=0.001$ and $p=0.8$, with a \textit{NDCG}@100 of 0.165. This demonstrates that the proposed augmentation strategy can preserve the item ranking order while providing more item correlation patterns, and thus improve the top-M performance. We then study the performance of NCAE on the \textit{HR}@100 metric, where "NCAE+" means NCAE with data augmentation, $\epsilon=0.001$ and $p=0.8$. As shown in Fig. \ref{fig:experi_data_aug}(b), we can see that the data-augmentation strategy can speed up convergence at the beginning of learning and finally get a better \textit{HR}@100. The relative improvement on \textit{NDCG}@100 (32.0\%) is much larger than on \textit{HR}@100 (4.2\%), indicating that NCAE+ can discover better ranking positions of true positive items during inference. 

Table \ref{tab:scale} shows the recommendation performance w.r.t users with different scale of interactions on Delicious. For instance, the second user cluster "$1$-$5(68)$" contains $68$ users with the number of ratings in $(1,5]$ (except $[0,1]$ on the first cluster). As can be seen, NCAE+ consistently outperforms NCAE, with the relative improvements of 127.1\%, 55.0\%, 36.4\%, 52.9\%, 19.6\% (averaged by two metrics) on the five clusters. The sparsity-aware data augmentation can improve  inactive user recommendation, since more item correlation patterns are provided for sparse users.


\mypar{2) Comparisons with Baselines}

To illustrate the effectiveness of our NCAE architecture, we propose a variant "NCAE-" that reweights the errors of all unobserved interactions uniformly with a confidence value $c=0.05$, which is the same to WMF. Specifically, we do not plot the results of NCAE- with $c=0$, since it is easy to overfit the training set and gives poor predictive performance without the error reweighting module. Besides, "NCAE+" means the utilization of data-augmentation. 

Fig. \ref{fig:experi_implict_1} and Fig. \ref{fig:experi_implict_2} show the performance of \textit{HR}@100 and \textit{NDCG}@100 on Delicious and Lastfm with respect to the number of hidden factors $K$. First, we can see that, with the increase of $K$, the performance of most models is greatly improved. This is because a larger $K$ generally enlarges the model capacity. However, large factors may cause overfitting and degrade the performance, e.g., eALS on Lastfm. Second, comparing two whole-based (uniform) methods NCAE- and WMF, we observe that by using non-linear matrix factorization, NCAE- consistently outperforms WMF by a large margin. The high expressiveness of neural network models is sufficient to capture subtle hidden factors when modeling the interaction data. The same conclusion can be obtained from the comparison of two sample-based methods NCF and BPR. Third, NCAE achieves the best performance on both datasets, significantly outperforming the strong baselines NCF and eALS. Specifically, the average improvements of NCAE+ over other models\footnote{We compute the relative improvements on both \textit{HR}@100 and \textit{NDCG}@100 metrics for all models except POP, since POP do not work on Delicious dataset. We report the average improvements of NCAE over all baselines and metrics.} are 170.4\%, 312.9\%, 170.0\%, 126.4\%, 111.0\% when $K$=8, 16, 32, 64, 128 on Delicious, verifying the effectiveness of our architecture and sparsity-aware data-augmentation. On Lastfm dataset, NCAE+ do not significantly outperforms NCAE, which is different to that on Delicious. We notice that there are two significant differences on the two datasets: 1) Lastfm is denser than Delicious; 2) Users of Lastfm are more preferred to listen the \textit{popular} artists, which can be observed from the excellent performance of POP method on Lastfm. Thus, our data-augmentation strategy that drops popular items may not work well on Lastfm. The average improvements of NCAE+ over other models are 70.5\%, 68.7\%, 72.4\%, 77.5\%, 78.1\% when $K$=8, 16, 32, 64, 128 on Lastfm.

Fig. \ref{fig:experi_implict_3} and Fig. \ref{fig:experi_implict_4} show the performance of Top-M recommendation, where $K$ is set to the best value for each model. First, we can see that eALS and NCF are strong baselines that beats BPR and WMF for all $M$, which is consistent with the results of \cite{he2016fast,he2017neural}. Second, neural network models (e.g., NCF and NCAE-) outperform other models, especially on \textit{NDCG}@M metrics of Delicious dataset, verifying the highly non-linear expressiveness of neural architectures. The performance of NCAE- can be further enhanced via utilizing the popularity-based error reweighting module (NCAE) and the sparsity-aware data augmentation (NCAE+). Lastly, NCAE+ consistently outperforms the best baseline NCF for every ranking position $M$, with the relative improvements of 33.0\%, 24.3\%, 22.3\%, 20.5\%, 18.1\% when $M$=50, 100, 150, 200, 250 on Delicious, and 41.5\%, 30.7\%, 29.7\%, 26.5\%, 28.1\% on Lastfm. Note that NCAE has fewer parameters than NCF (NCF stacks three hidden layers while NCAE is a shallow network). The autoencoder-based structures that directly take user vectors as inputs for batch training, may be better for collaborative filtering (the architecture of NCF is designed at the interaction level). Generally, the smaller $M$ is, the larger improvement of NCAE+ against other models, indicating the reliability of our model on top-M recommendation.


\subsection{Robustness and Scalability}
\mypar{1) Analysis of Robustness}
\begin{table}[t]
  \setlength{\abovecaptionskip}{3pt}
  \setlength{\belowcaptionskip}{0pt}
\centering
\caption{Performance on Lastfm with varying data ratio.}
\label{tab:robustness}
\begin{tabular}{clccc}
\hline
Algorithms & Metrics & $D=40\%$ & $D=60\%$ & $D=80\%$\\
\hline
\multirow{4}*{eALS} & \textit{\small{HR@50}} & 0.082 & 0.107 & 0.119\\
&\textit{\small{NDCG@50}}& 0.023 & 0.030 & 0.038\\
&\textit{\small{HR@100}} & 0.132 & 0.165 & 0.186\\
&\textit{\small{NDCG@100}} & 0.030 & 0.041 & 0.047\\
\hline
\multirow{4}*{NCF} 
&\textit{\small{HR@50}} & 0.101 & 0.125 & 0.147\\
&\textit{\small{NDCG@50}}& 0.030 & 0.040 & 0.047\\
&\textit{\small{HR@100}} & 0.154 & 0.174 & 0.195\\
&\textit{\small{NDCG@100}} & 0.040 & 0.050 &0.056\\
\hline
\multirow{4}*{NCAE+} & \textit{\small{HR@50}} & 0.123 & 0.155 & 0.167\\
&\textit{\small{NDCG@50}}& 0.042 & 0.054 & 0.061\\
&\textit{\small{HR@100}} & 0.176 & 0.199 & 0.222\\
&\textit{\small{NDCG@100}} & 0.050 & 0.062 &0.071\\
\hline
\end{tabular}
\end{table}

We now explore the robustness of eALS, NCF and NCAE+ to different proportions of training set exploited. We randomly sample a $D$ proportion of interaction data from each user. The test set and the hyperparameters remain unchanged. As shown in Table \ref{tab:robustness}, we can observe that NCAE+ significantly outperforms the strong baselines eALS and NCF over all ranges of $D$, and NCAE+ (40\%) has already performed slightly better than eALS (80\%) and NCF (60\%). Specifically, the average improvements of NCAE+ over eALS are 50.4\%, 62.4\%, 66.3\% when training density $D$=80\%, 60\%, 40\% on top-50 metrics, and 35.2\%, 35.9\%, 50.0\% on top-100 metrics. Generally, as the training set gets sparser, NCAE+ gets a larger improvement against eALS (or NCF). This demonstrates the robustness of our model under the sparsity scenario.

\mypar{2) Scalability and Computation Time}

Analytically, by utilizing the sparse forward module and the sparse backward module\footnote{We use \textit{tf.sparse\_tensor\_dense\_matmul()} and \textit{tf.gather\_nd()} to implement NCAE.}, the training complexity for one user $i$ is $O(K_1|\Ri|+\sum_{l=1}^{L-2}K_{l+1}K_{l}+|\Ri|K_{L-1})$, where $L$ is the number of layers and $K_l$ is the hidden dimension of layer $l$. All our experiments are conducted with one NVIDIA Tesla K40C GPU. For Lastfm and Delicious datasets, each epoch takes only 0.4 seconds and 2.1 seconds respectively, and each run takes 30 epochs. 

For ML-10M dataset, it takes about 6.3 seconds and needs 60 epochs to get satisfactory performance. Since ML-10M is much larger than the other two datasets, this shows that NCAE is very scalable. Table \ref{tab:scalability} shows the average training time per epoch on ML-10M dataset w.r.t. hidden factors $K$ and training set ratio $D$, where a 3-layer item-based NCAE with equal hidden factors is used. Then the time complexity can be reformulated as $O(K|\Rj|+K^2)$. As can be seen, the runtime scales almost linearly with $K$, since the modern GPU architecture parallelizes the matrix-vector product operation automatically. Besides, further observation shows that, the runtime grows slowly as the training size $D$ increases, indicating the scalability of NCAE on large datasets. In practice, the second term $O(\sum_{l=1}^{L-2}K_{l+1}K_{l})$ overtakes the extra cost.
\begin{table}[t]
  \setlength{\abovecaptionskip}{3pt}
  \setlength{\belowcaptionskip}{0pt}
\centering
\caption{Training time per epoch on ML-10M (in seconds).}
\label{tab:scalability}
\begin{tabular}{ccccc}
\hline
\quad $K$ & $D=10\%$ & $D=30\%$ & $D=50\%$ & $D=100\%$\\
\hline
\quad 100 & 1.31 & 2.34 & 3.22 & 4.95\\
\quad 300 & 1.99 & 3.10 & 3.98 & 5.68\\
\quad 500 & 2.48 & 3.85 & 4.68 & 6.31\\
\hline
\end{tabular}
\end{table}

 
\section{Conclusion and Future Work}
\label{sec:conclusions}
In this paper, we present a new framework named NCAE for both explicit feedback and implicit feedback, and adapt several effective deep learning approaches to the recommendation domain. Contrary to other attempts with neural networks, we employ a new loss function for sparse inputs, and propose a three-stage pre-training mechanism, an error reweighting module and a data-augmentation strategy to enhance the recommendation performance. We conducted a comprehensive set of experiments to study the impacts of difference components, verifying the effectiveness of our NCAE architecture. We also compared NCAE with several state-of-the-art methods on both explicit and implicit settings. The results show that NCAE consistently outperforms other methods by a large margin.

In future, we will deal with the cold-start problem and explore how to incorporate auxiliary information. We also plan to study the effectiveness of neural network models when an online protocol is used \cite{he2016fast}. Another promising direction is to explore the potential of other neural structures for recommendation. Besides, it is important to design an interpretative deep model that can explicitly capture changes in user interest or item properties.

\bibliographystyle{abbrv}
\bibliography{tkde_ncae.bib}
\ifCLASSOPTIONcaptionsoff
  \newpage
\fi

\end{document}